\documentclass[12pt]{article}

\usepackage{multicol}
\usepackage{amssymb} % The amssymb package provides various useful mathematical symbols
\usepackage{amsmath} % Math functions
\usepackage{hyperref} % To use \href 
\usepackage{url} % To use \url
\usepackage[ddmmyyyy, hhmmss, 24hr]{datetime}
\usepackage{graphicx} % image inserting
\usepackage[font={small,sf}]{caption}
\usepackage[font={small,sf}]{subcaption}
\usepackage{siunitx} % IS units print
\usepackage{enumitem} % For lists
\usepackage{multirow}
\usepackage{longtable} % Multiple page table
\usepackage{array} % For fixed width in tabular
\usepackage{booktabs}
\usepackage{fancyhdr}

\usepackage[sorting=none, style=numeric, alldates=iso, alltimes=24h]{biblatex} % Bibliography management

\bibliography{BIBLIOGRAPHY} % Load bib file also with \addbibresource{<database>.<extension>}

\begin{document}

\pagestyle{fancy}
\fancyhf{}
\lhead{ \fancyplain{}{Lafiosca et al.} }
\rhead{ \fancyplain{}{\textit{Towards a New Model for Dent Evaluation}} }
\cfoot{ \fancyplain{}{\thepage} }

% Title

\title{Aircraft Skin Inspections: Towards a New Model for Dent Evaluation} %Declares the document's title.
%\subtitle{Subtitle subtitle subtitle subtitle subtitle subtitle subtitle}

%Authors, affiliations address.`
\renewcommand{\thefootnote}{\arabic{footnote}}

%Author with Email%
\author{Pasquale Lafiosca\thanks{Corresponding author: pasquale.lafiosca@cranfield.ac.uk ORCID 0000-0002-3396-5744 }, Ip-Shing Fan\thanks{ORCID 0000-0001-6691-935X} and Nicolas P. Avdelidis\thanks{ORCID 0000-0003-1314-0603} \\
    \textit{Integrated Vehicle Health Management Centre} \\
    \textit{Cranfield University} \\
    \textit{United Kingdom}
    }

\date{First version: 16/05/2022 \\ Last revision: 06/03/2023}

\maketitle                   % Produces the title.

%Abstract
\begin{abstract}
Aircraft maintenance, repair and overhaul industry is gradually switching to 3D scanning for dent inspection.
High-accuracy devices allow quick and repeatable measurements, which translate into efficient reporting and more objective damage evaluations.
However, the potential of 3D scanners is far from being exploited. This is due to the traditional way in which the structural repair manual deals with dents, that is, considering length, width and depth as the only relevant measures. Being equivalent to describing a dent similarly to a ``box'', the current approach discards any information about the actual shape.
This causes high degrees of ambiguity, with very different shapes (and corresponding fatigue life) being classified as the same, and nullifies the effort of acquiring such great amount of information from high-accuracy 3D scanners.
In this paper a $7$-parameter model is proposed to describe the actual dent shape, thus enabling the exploitation of the high fidelity data produced by 3D scanners.
The compact set of values can then be compared against historical data and structural evaluations based on the same model. 
The proposed approach has been evaluated in both simulations and point cloud data generated by 8tree's dentCHECK tool, suggesting increased capability to evaluate damage, enabling more targeted interventions and, ultimately, saving costs.

\subsubsection*{Keywords}
3D scanning; Aircraft inspections; Dents; Maintenance; Point clouds.
    
\end{abstract}

\setcounter{footnote}{0}
%% main text
\section{INTRODUCTION}\label{sec:DentIntro}

Aircraft skin is generally built with aluminium alloys or fibre-reinforced composites. 
Regardless of the differences in weight, fatigue performance, corrosion resistance and other material properties, the skin is exposed to various types of damage~\cite{dursun2014recent, VisualInspectionReliabilityForCompositeAircraftStructures, chen2014inspection}.

% Follow SRM indications or follow instructions of the aircraft manufacturer
The \textit{structural repair manual} (SRM) is the main document containing information to identify and repair damages on primary and secondary structures. The former include all those structures that contribute significantly to carrying flight, ground and pressurisation loads, with most of the skin usually falling into this category, as for example stated in the Boeing 737-400 SRM~\cite{SRM737400}.

\textit{Dents}, in particular, are extremely common during operations due to accidental impact with service vehicles, hail, small rocks on the runway or any other object that may cause damage, also known as \textit{foreign object debris} (FOD).
% Why dent are important
They affect structural performance of the thin aircraft skin, shortening fatigue life and contributing in the propagation of cracks~\cite{li2017experimental,zhang2010influence,shivalli2006effect}.
% How dent are found and measured (subjectivity is not the issue here, lack of model is!)
For this reason, during maintenance tasks, the skin is thoroughly inspected, seeking for dents among many other anomalies that must be carefully evaluated.

\begin{figure}[h!]
 \centering
    \includegraphics[width=0.5\linewidth]{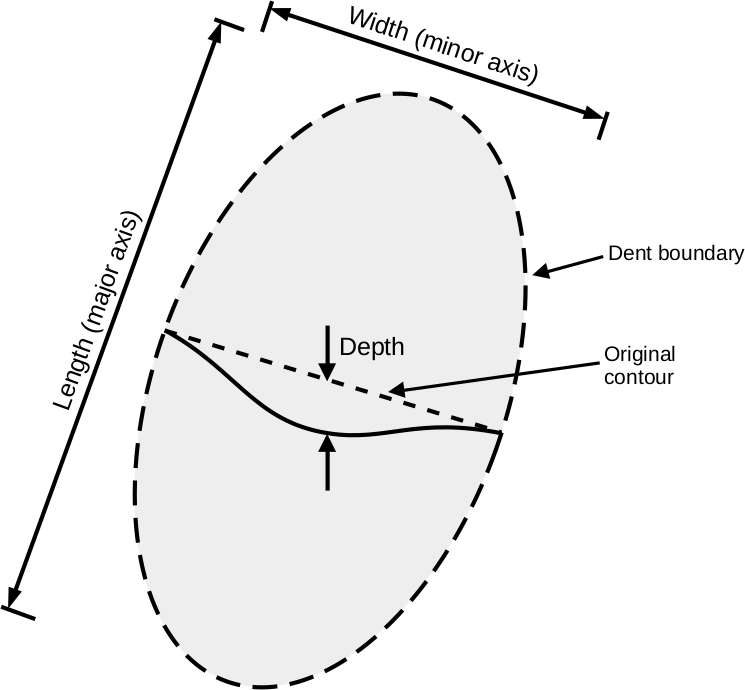}
  \caption{Traditional dimensional evaluation of dents.}
  \label{fig:dentScheme}
\end{figure}

The Boeing 737-400 SRM~\cite{SRM737400}, here taken as reference, provides the only accepted definitions for such type of aircraft. A dent on metal is \textit{``a damaged area that is pushed in from its normal contour with no change in the cross-sectional area of the material. The edges of the damaged area are smooth. This damage is usually caused by a hit from a smoothly contoured object. The length of the dent is the
longest distance from one end to the other end. The width of the dent is the second longest
distance across the dent, measured at 90 degrees to the direction of the length''}~\cite{SRM737400} (see Fig.~\ref{fig:dentScheme}). The depth is measured where the width is taken and there must be no other kind of damages like gouges or cracks at the same location~\cite{SRM737400}.
Similarly, a dent on composites is \textit{a concave depression which does not break the fibers}~\cite{SRM737400}.

\begin{figure}[h!]
 \centering
  \begin{subfigure}[t]{0.4\linewidth}
  \centering
    \includegraphics[width=\linewidth]{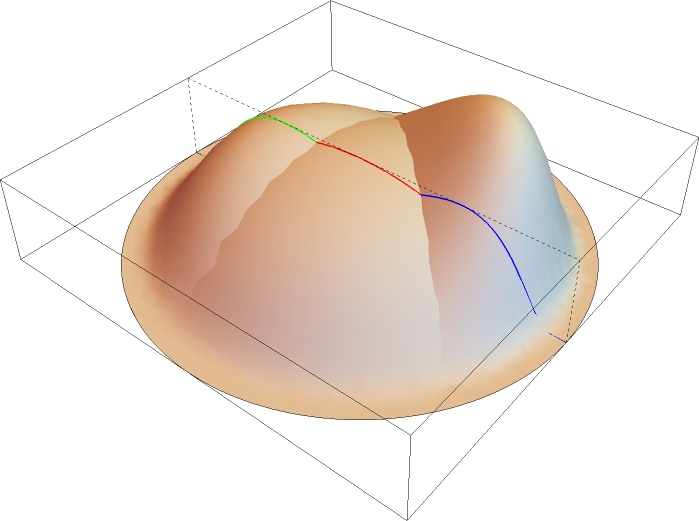}
    \caption{Three very different dents, but falling as the same according to the SRM.}
  \end{subfigure}
  \hspace{0.1\linewidth}
  \begin{subfigure}[t]{0.4\linewidth}
   \centering
    \includegraphics[width=\linewidth]{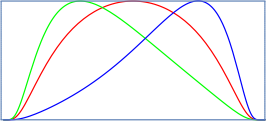}
    \caption{Section where the width and the height should be measured.}
  \end{subfigure}
  
  \caption{SRM definition implicitly model a dent like a ``box''.}
  \label{fig:dentBox}
\end{figure}

During maintenance, repair and overhaul (MRO) operations dents must be \textit{found}, \textit{measured} and \textit{evaluated} in order to identify the necessary repair actions, if any~\cite{breuer2016commercial}.
The above definition implicitly defines the currently used model for dents as a ``box'', whose dimensions are equal to the length, width and depth of the dent. Moreover, it must be remarked that, as the depth is measured where the width is taken, the box might not even be the one completely containing the dent.
Fig.~\ref{fig:dentBox} shows a set of dents falling within the same box.

Research has shown that residual strength is connected not only to dent amplitude (and its position), but also to the form of the deformation~\cite{dow1984effects, li2015effect}.
The traditional 3 parameters are not sufficiently representative of the complexity of dent damage: in particular, they cannot properly represent dents where the boundary is irregular or when the deepest point is not centred, or express how quickly the maximum depth is reached.
In the event that the appropriate repair procedure is not contained in the SRM, the repair instructions must be obtained from the aircraft manufacturer. In this case, the need of a reliable damage representation is even more evident.

This work addresses the clear need for a more accurate model for dent evaluation in the current context of MRO, which enables the full exploitation of new 3D scanning devices, originally introduced to improve inspection and maintenance processes, but still seeing limited use.

\subsection{Similar works}

To date, methods to characterise and report the shape of a dent can only be found in technical documentation issued by aircraft manufacturers, containing practices addressed to maintenance engineers and sharing characteristics similar to the ones described in this paper.
The academic literature on aircraft dents is mostly related to their structural evaluation.
The compressive failure load that a dented aluminium panel can sustain was found generally lower compared to the virgin panel~\cite{lang2007investigation}, but dents were reported only as depth and width.
Similarly, Zhigang et al.~\cite{li2017experimental} showed how the residual strength of aluminium specimens is significantly affected by dents.
The inspection and report of dents is a costly and time consuming process and previous research has shown how the optimisation of the inspection intervals may reduce costs~\cite{jing2021inspection}.

Dent characterisation can also be found in the literature regarding pipeline engineering. Dinovitzer et al.~\cite{dinovitzer2002geometric} presented some relations between service life and dent geometry, showing that depth alone cannot describe how likely a failure is. Dent depth was also related to the critical pressure of a pipeline~\cite{allouti2012study}.
Recently, aviation MRO, as well as other industries, has been reached by computer-aided dent inspections, made possible by the use of 3D scanning systems~\cite{8treeAMC2022, allard2013improvement}. Despite the accurate 3D information that can be extracted, these advanced tools have not addressed the problem of having a compact model to describe the dent shape.

\section{PROBLEM STATEMENT}\label{sec:problemStatement}
As anticipated above, the SRM first classifies a dent by its \textit{dimensional} characteristics, namely length, width and depth.
Apart from other properties, like the type of material and the position with respect to certain aircraft parts or other dents, the measures of width and width/depth ratio are the only dimensional ones used to classify the damage as \textit{allowable} or \textit{not allowable}, the latter meaning that a repair is needed, immediately or within a certain number of cycles (if some conditions are met).
This information is then reported in the \textit{dent and buckle chart}. Often a representation of the damage is sent to the aircraft manufacturer, who provides detailed instructions about how to conduct the repair.

Despite the dimensional characteristics being paramount for a proper evaluation, their reliability is undermined by many factors~\cite{VisualInspectionReliabilityForCompositeAircraftStructures}.
In fact, in the vast majority of cases, measures are collected manually by engineers by using a depth gauge and a ruler, thus a certain degree of subjectivity and error is always present.
To introduce measurement repeatability, 3D scanning devices are being gradually introduced in MRO. These devices  produce a point cloud representation of the scanned surface and help the engineer producing a report.
For example, in a study carried out by the company \textit{8tree} the average standard deviation in width/depth ratio was $23.2$ for traditional measuring and only $1.3$ when using their 3D scanning device~\cite{8tree100Measures}.

% What is the problem of NOT having a model
Nevertheless, 3D scanning does not entirely solve the problem: while the subjectivity of the measure is eliminated and the damage representation is moved from the physical to the digital world as 3D point cloud, most of the shape information is lost when comparing against the simplistic dimensional description and evaluation procedure prescribed by the SRM.
In other words, despite the remarkable accuracy reached by 3D scanners for MRO (currently in the range of $\SI{25}{\micro\meter}$), the assessment of the damage follows a \textit{``box'' model} that does not take into account the actual shape. Moreover the final evaluation is typically made according to width and width/depth ratio, nullifying the effort to acquire such great level of detail.

% What is a model
Indeed, a model must provide a simplified representation of reality, that does not include all the real attributes, but only those that are \textit{relevant} to the issue under consideration, within a given tolerance~\cite{generalModelTheory}. 
The current SRM dent representation is ambiguous and insufficient for an objective evaluation of the damage, especially with respect to the amount of information available by 3D scanning.

\begin{figure}[!ht]
 \centering
  
  \begin{subfigure}[t]{0.3\linewidth}
  \centering
    \includegraphics[width=\linewidth]{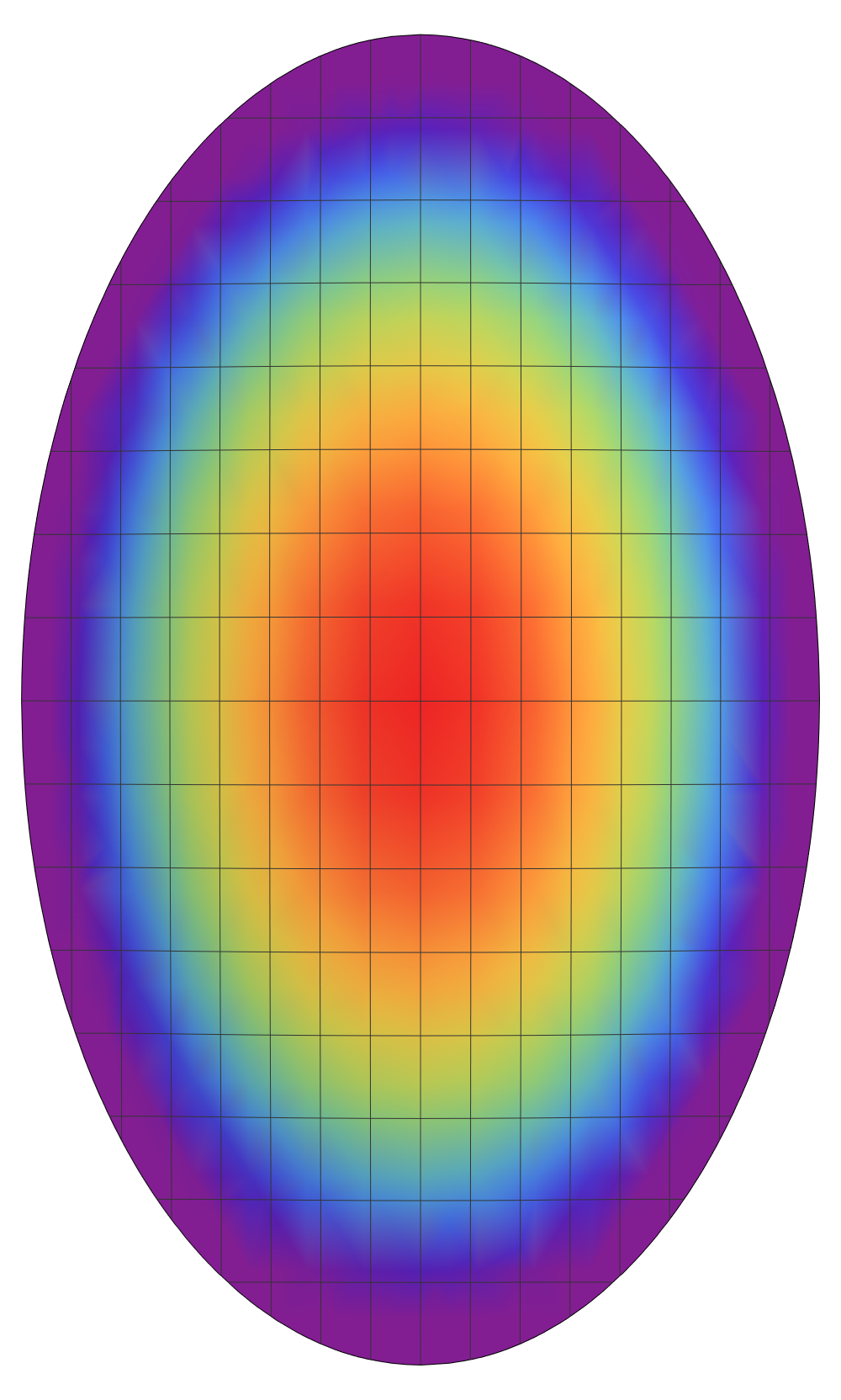}
  \end{subfigure}
  %\hspace{0.05\linewidth}
  \begin{subfigure}[t]{0.3\linewidth}
   \centering
    \includegraphics[width=\linewidth]{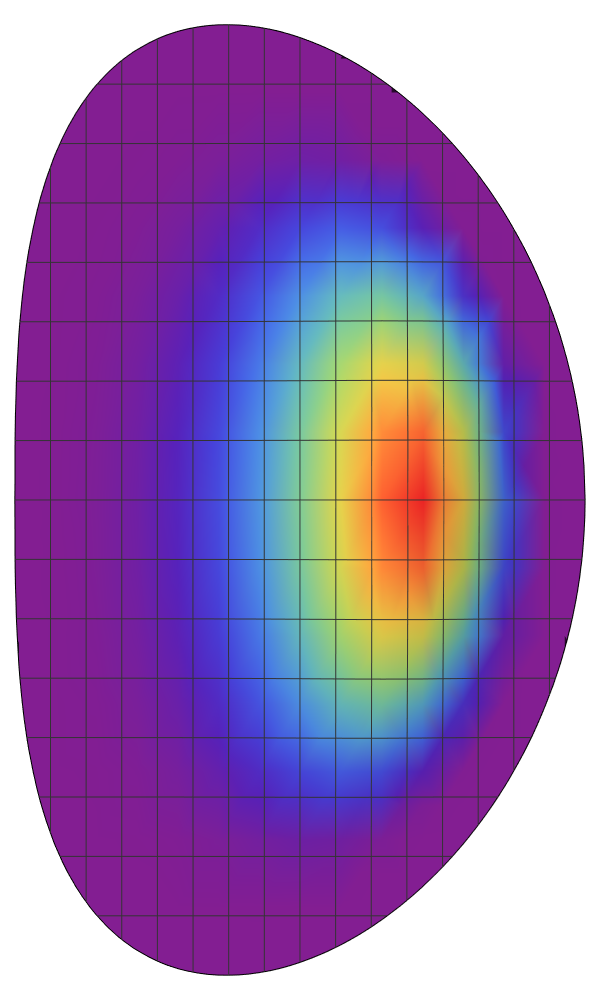}
  \end{subfigure}
  %\hspace{0.05\linewidth}
  \begin{subfigure}[t]{0.3\linewidth}
   \centering
    \includegraphics[width=\linewidth]{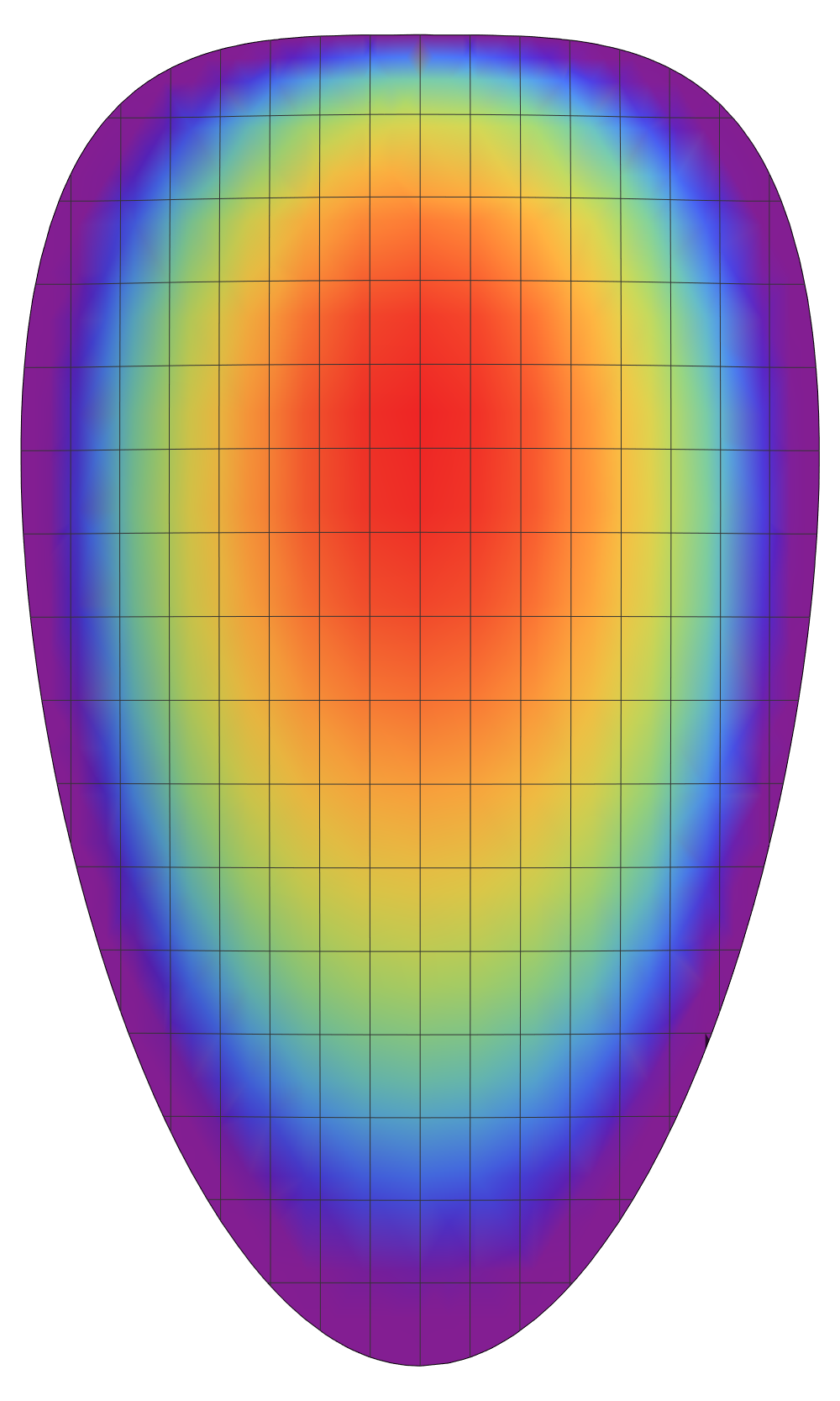}
  \end{subfigure}

\caption{These three illustrative dents have all the same length and width and depth (as heatmap) and are thus classified in the same way by the SRM. However, they have evidently different shapes.}
\label{fig:illustrativeDents}
\end{figure}

Consider, for example, the three illustrative dents represented in Fig.~\ref{fig:illustrativeDents}.
By following the SRM to the letter, these dents should be classified in the exact same way, yet they have completely different shapes. In all likelihood, also their structural characteristics are different.

% Why model is important
Hence it follows the need for a more accurate model, against which tailored structural evaluations may be carried out, redefining the allowable damage thresholds accordingly. In practice, this translates into avoiding unnecessary repairs and, ultimately, reducing costs. To the best of our knowledge, no alternative dent damage representation and evaluation strategy is present in the literature.

% Hook for next section...
The next section proposes a mathematical model for the shape of a generic dent. When coupled with 3D scanning, it paves the way towards a new benchmark for dent classification and evaluation.

\section{PROPOSED MODEL}\label{sec:DentModel}

\subsection{Definition}
The model here reported is an elaboration of the so-called \textit{bump} function~\cite{johnson2015saddle}, having characteristics ascribable to a dent: smooth shape and vanishing boundaries.
While a simpler derivation has been used to conduct simulations and build specimens~\cite{lafioscaDentSegmentation}, it does not have the capability to resemble asymmetric and irregular shapes. The general model here presented, instead, introduces \textit{egg-like} boundaries, \textit{xy}-shifted maximal depth and variable exponential base for a better representation of the dimensional characteristics, providing $4$ extra parameters in addition to the traditional $3$ (length, width and depth) already employed in the SRM.

The proposed dent model is first introduced by defining a \textit{reference} dent:
\begin{equation}
%\resizebox{\linewidth}{!}{
%$
\operatorname{refDent}(x,y) \triangleq
  \begin{cases}
  
    % 1st
    b^{ \Big[ 1 - \frac{1}{1-\mathrm{r}^2(x,y)} \Big] } , & 
    
    \begin{aligned}
     & -f(x)<y<f(x)  \\
     & \wedge s_x=0 \wedge s_y=0
    \end{aligned} \\[1.5em]

    % 2nd
    \begin{aligned}
    b^{ \Big[ -\frac{1}{1-\mathrm{r}^2(x,y)} + (x-s_x)\frac{2 r(s_x,s_y) \frac{\partial r}{\partial x}(s_x,s_y) }{(1-\mathrm{r}  ^2(x,y))^2}} \\
    ^{(y-s_y)\frac{2 r(s_x,s_y) \frac{\partial r}{\partial y}(s_x,s_y) }{(1-\mathrm{r}^2(x,y))^2}
     + \frac{1}{1-\mathrm{r}^2(s_x,s_y)} \Big] } 
    \end{aligned} , &

    \begin{aligned}
     & -f(x)<y<f(x)  \\
     & \wedge (s_x\neq0 \vee s_y\neq0)
    \end{aligned} \\[1.5em]
    
    % 3rd
    \text{undefined}, & \text{elsewhere}
  
  \end{cases}
%$
%}
\end{equation}

\begin{figure}[!h]
 \centering
 \begin{subfigure}[b]{0.45\linewidth}
 \centering
    \includegraphics[width=\linewidth]{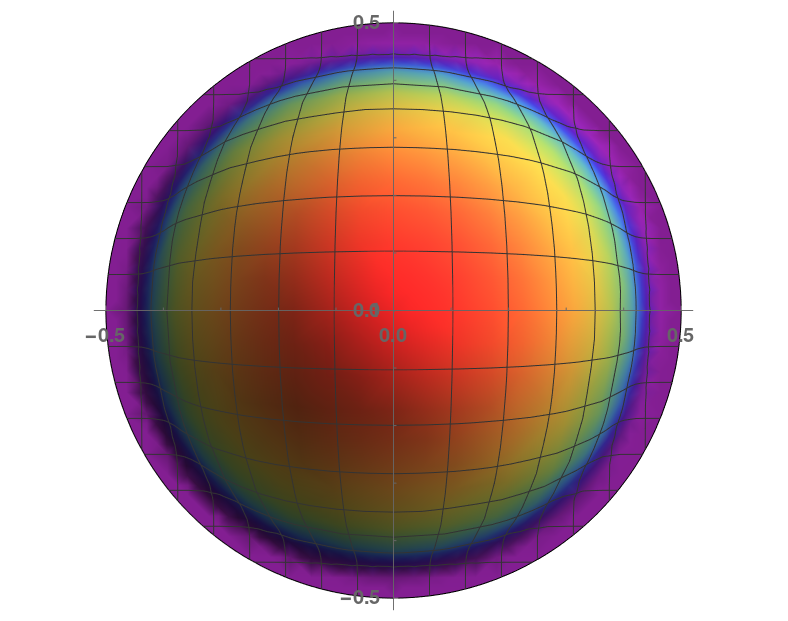}
    \caption{Top view.}
  \end{subfigure}
  %\hspace{0.1\linewidth}
  \begin{subfigure}[b]{0.45\linewidth}
  \centering
    \includegraphics[width=\linewidth]{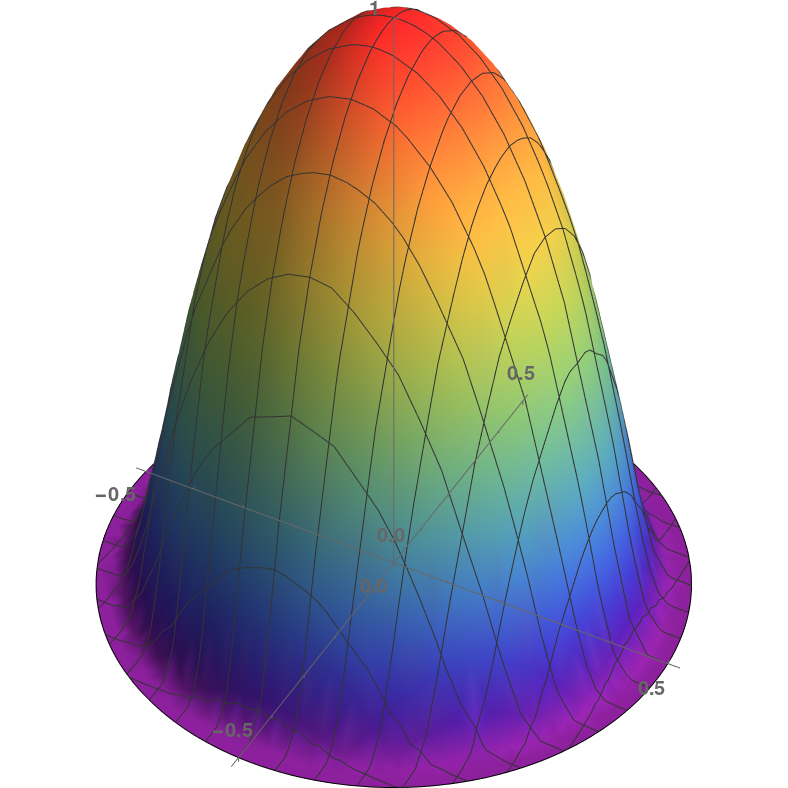}
    \caption{General view.}
  \end{subfigure}
  %\caption*{\rightline{\scriptsize{Credits: .} }}
  \caption{Reference dent function $\operatorname{refDent}(x,y)$ with unitary $xyz$ dimensions, $p=1$, $s_x = s_y = 0$ and $b = e$.}
  \label{fig:dentReference}
\end{figure}

The depth is always zero along the boundary, beyond which the dent ceases to be such and the function is considered undefined. The helper function $r(x,y)$ is defined as:
\begin{equation}
    r(x,y) \triangleq \frac{\sqrt{x^2 + y^2}}{\sqrt{x^2 + f^2(x)}}
\end{equation}
with $\frac{\partial r}{\partial x}(x,y)$ and $\frac{\partial r}{\partial y}(x,y)$ its partial derivatives and:
\begin{equation}
    f(x) \triangleq \sqrt{0.25 -((x + 0.5)^p - 0.5)^2}
\end{equation}
The following $4$ extra parameters are thus employed to characterise a dent:
\begin{itemize}
 \item The base $b \in (1,\infty)$ is responsible for how fast the depth increases, going inward. Faster for values close to $1$;
 \item The \textit{egg-factor} $p \in (0,2)$ produces the egg-shaped boundary and it is adapted from the work of T.C. Carter~\cite{carter1968hen}. Note that for $p=1$ the boundary is a circumference and its behaviour is not symmetric between lower and greater values;
 \item $s_x \in (-0.5, 0.5)$ is the shift of the deepest point along the $x$-axis;
 \item $s_y \in (-0.5, 0.5)$ is the shift of the deepest point along the $y$-axis.
\end{itemize}

The reference function $\operatorname{refDent}(x,y)$ showed in Fig.~\ref{fig:dentReference} has unitary dimensions along the three axis and can be easily rescaled to produce a general dent:

\begin{equation}\label{eq:dent}
\operatorname{dent}(x,y) \triangleq d \cdot \operatorname{refDent}\left(\frac{x}{l},\frac{y}{w}\right)
\end{equation}
where $l$, $w$ and $d$ are the length, width and depth of the dent, respectively. Here the terms \textit{length} and \textit{width} are used for the dimensions along the $x$-axis and $y$-axis, respectively, and may be swapped with respect to the SRM definition (where the length is always the \textit{longest} dimension). It must be remarked that the depth is no longer measured in correspondence of the wider point, but mirrors the actual depth of the dent, thus resolving a major inconsistency of the traditional box model.

The above model assumes that a dent sits on a plane. In case of dents laying on leading edges or other complex surfaces, additional degrees of freedom must be considered to extract the dent from the undamaged surface.

\subsection{Examples}
Before discussing the possible applications in the following section, some shapes obtained with the proposed model are shown in Table~\ref{tab:examples}, accompanied by their corresponding parameters. Depth values have been intentionally exaggerated for visualisation purposes.

% Images
\begin{center}
\begin{longtable}{ m{0.35\linewidth} m{0.35\linewidth} p{0.2\linewidth} }
\toprule
\multicolumn{1}{c}{\textbf{Top view}} & \multicolumn{1}{c}{\textbf{Overall view}} &\multicolumn{1}{l}{\textbf{Parameters}}\\
\midrule
\endhead % all the lines above this will be repeated on every page

\includegraphics[width=\linewidth]{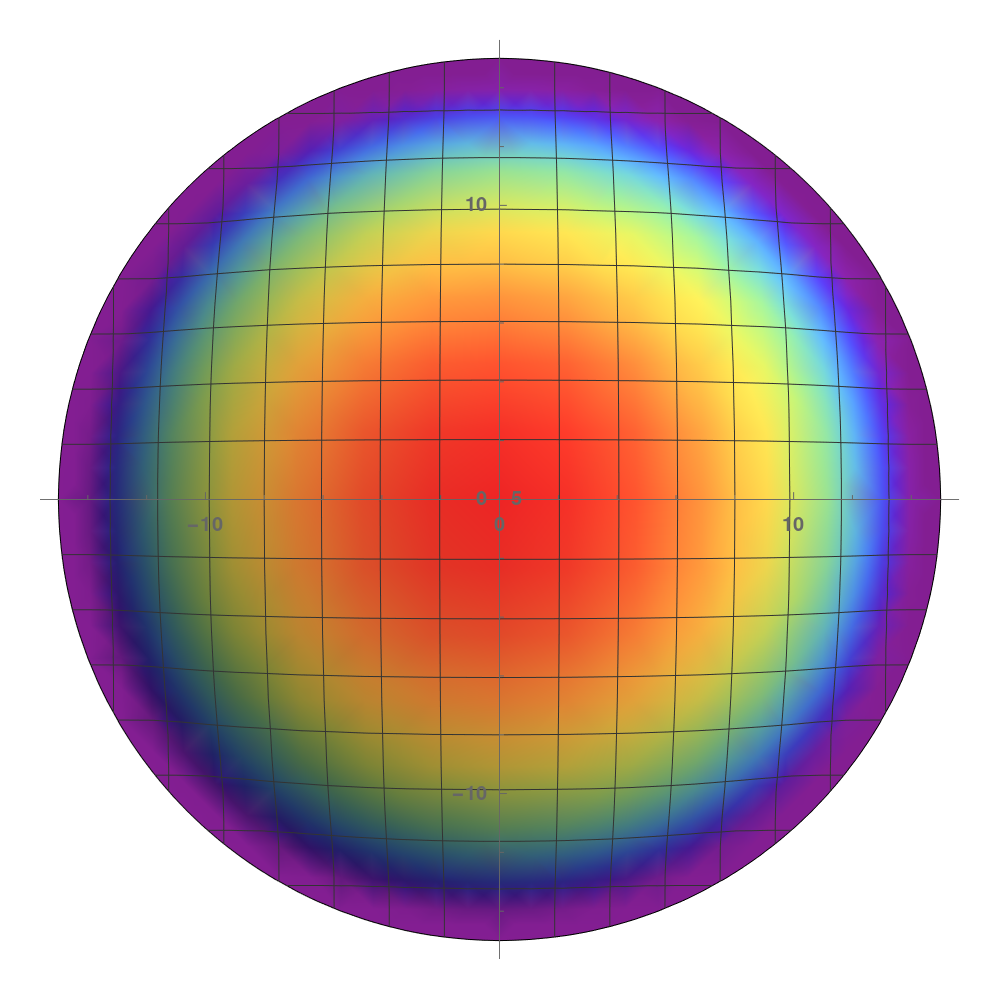} & \includegraphics[width=\linewidth]{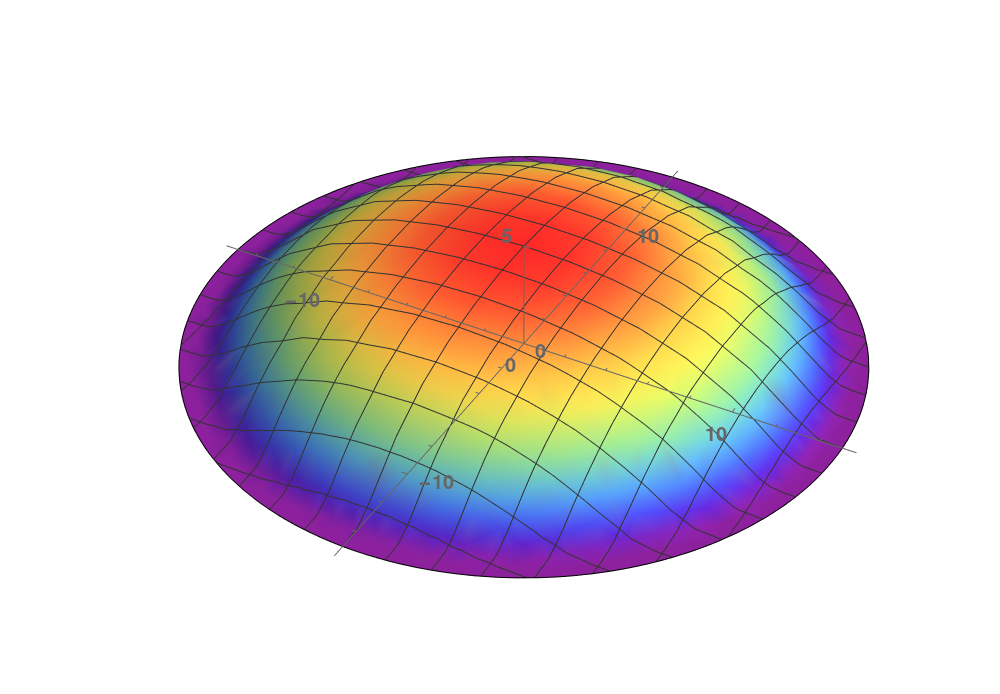}     &
\parbox{0.2\linewidth}{
 \begin{align*}
 l &= 30 \\
w &= 30 \\
d &= 5 \\
b &= 2 \\
p &= 1 \\
s_x &= 0 \\
s_y &= 0 \\
\end{align*}
} \\

\midrule
\includegraphics[width=\linewidth]{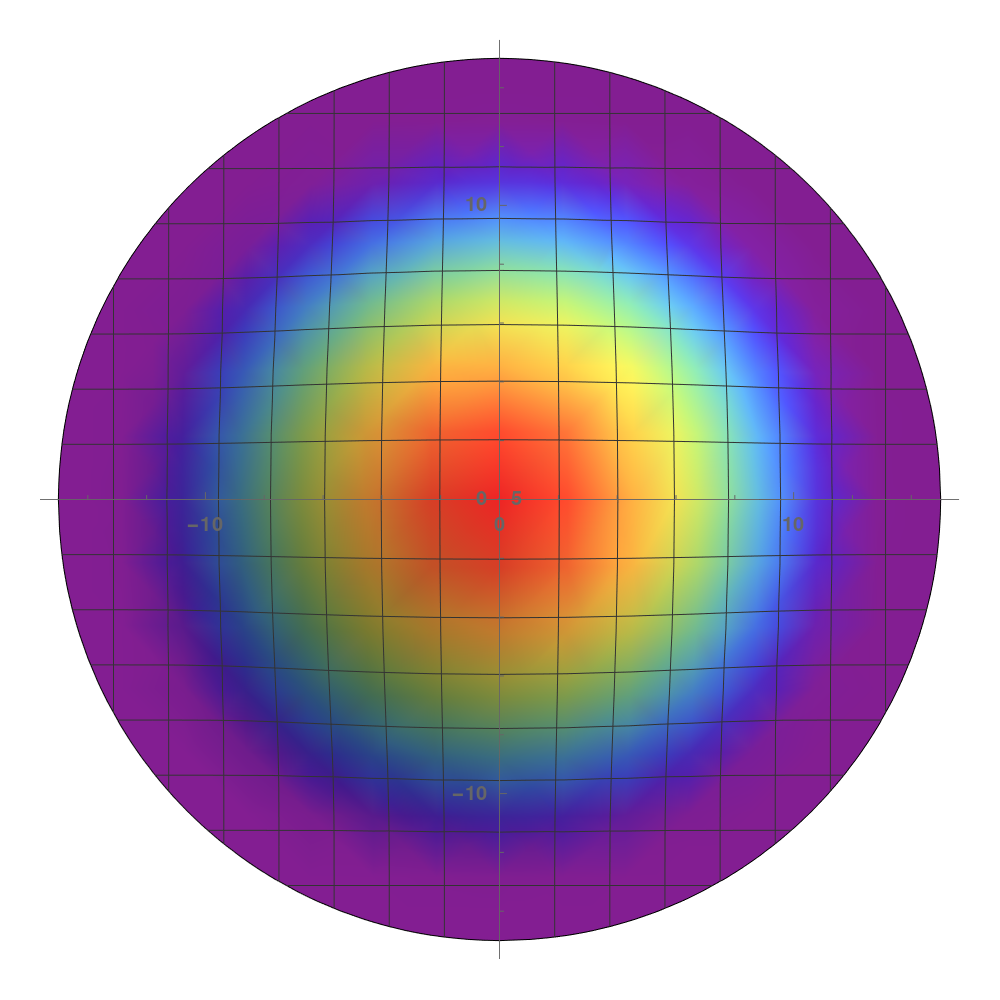} & \includegraphics[width=\linewidth]{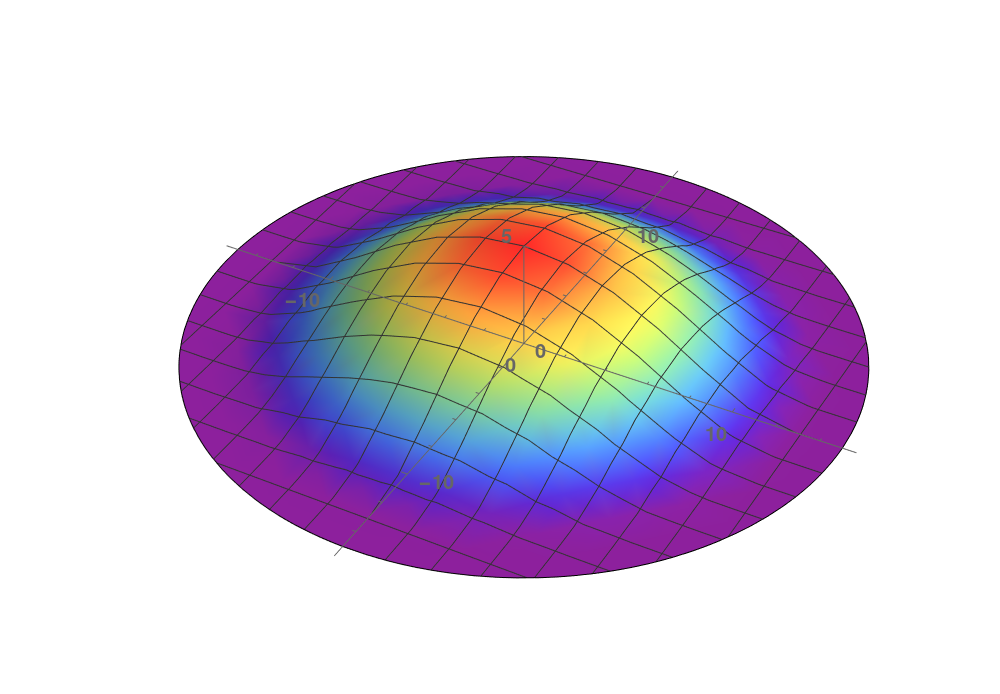}     &
\parbox{0.2\linewidth}{
 \begin{align*}
 l &= 30 \\
w &= 30 \\
d &= 5 \\
b &= 10 \\
p &= 1 \\
s_x &= 0 \\
s_y &= 0 \\
\end{align*}
} \\

\midrule
\includegraphics[width=\linewidth]{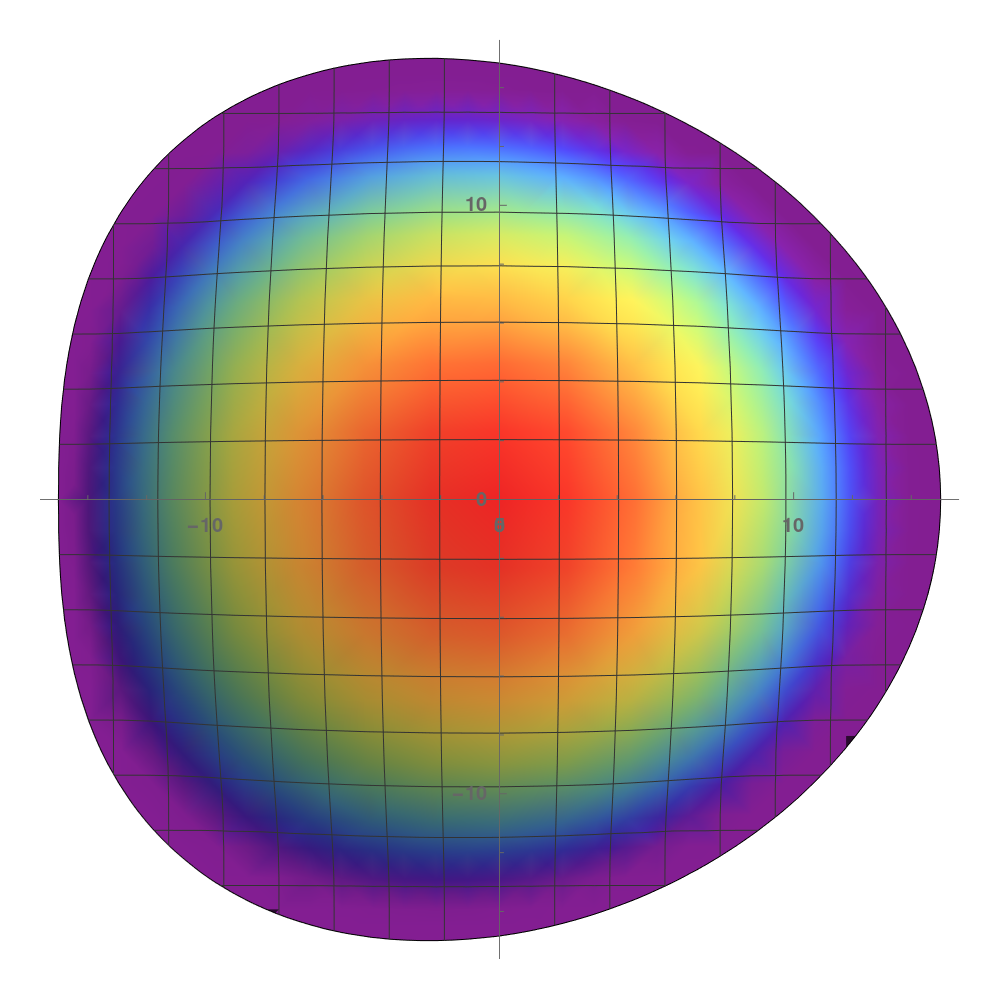} & \includegraphics[width=\linewidth]{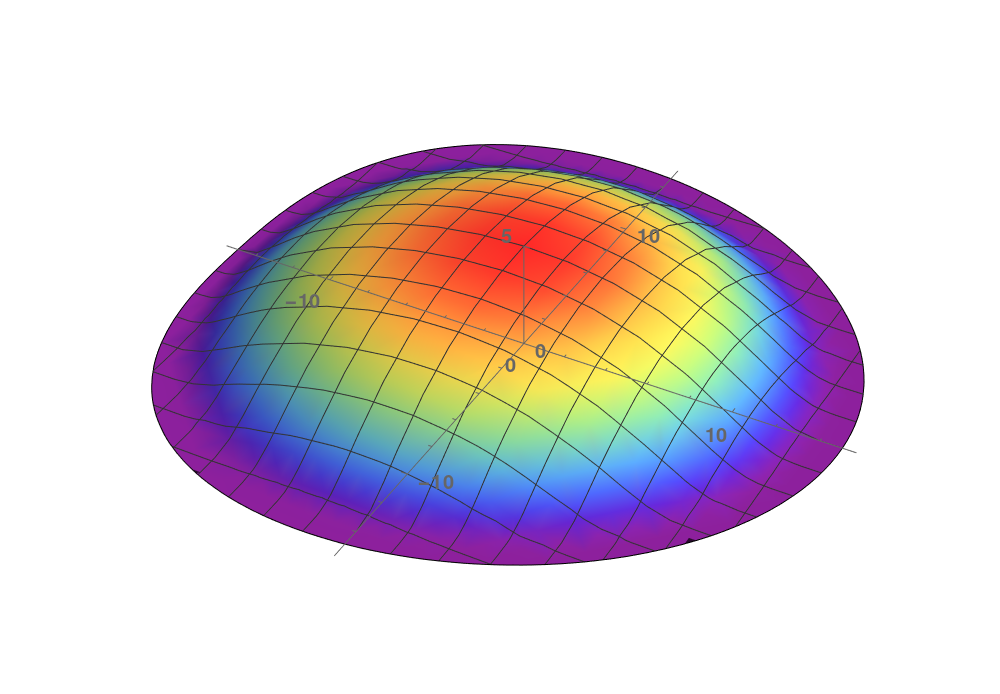}     &
\parbox{0.2\linewidth}{
 \begin{align*}
 l &= 30 \\
w &= 30 \\
d &= 5 \\
b &= e \\
p &= 0.8 \\
s_x &= 0 \\
s_y &= 0 \\
\end{align*}
} \\

\midrule
\includegraphics[width=\linewidth]{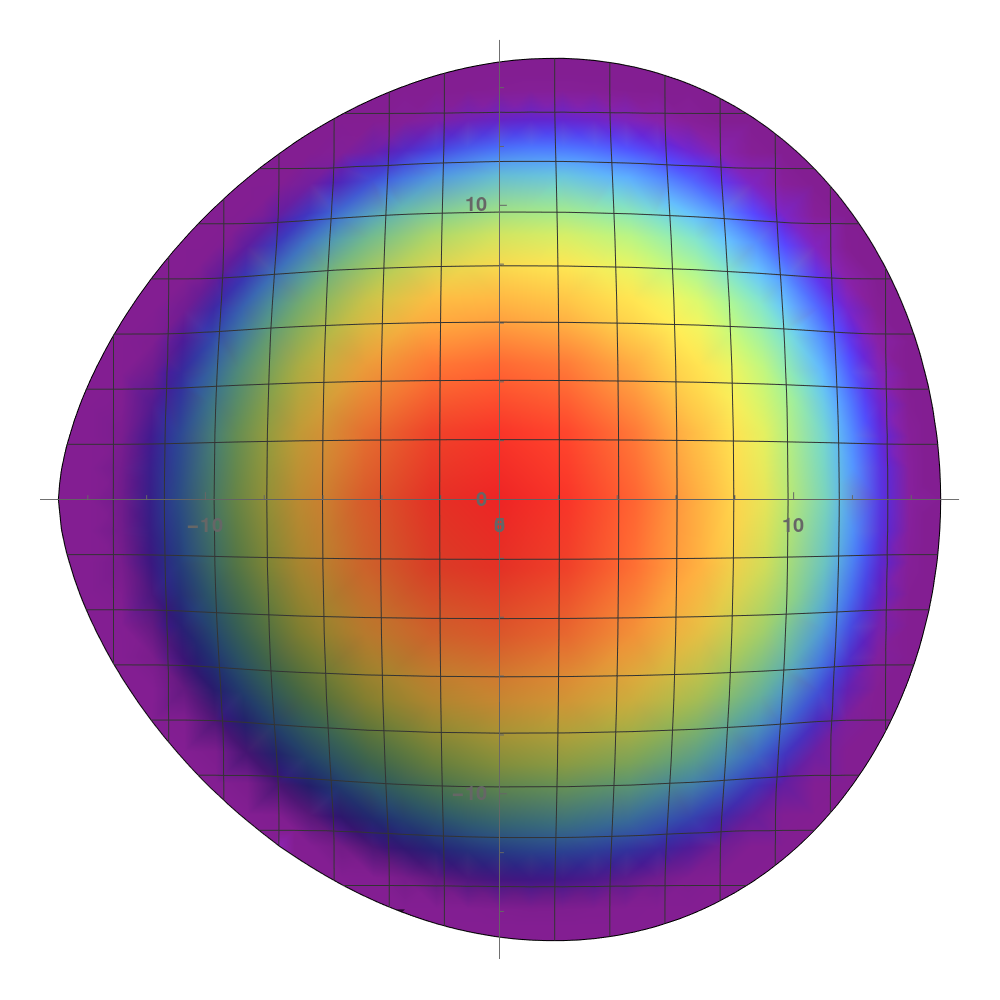} & \includegraphics[width=\linewidth]{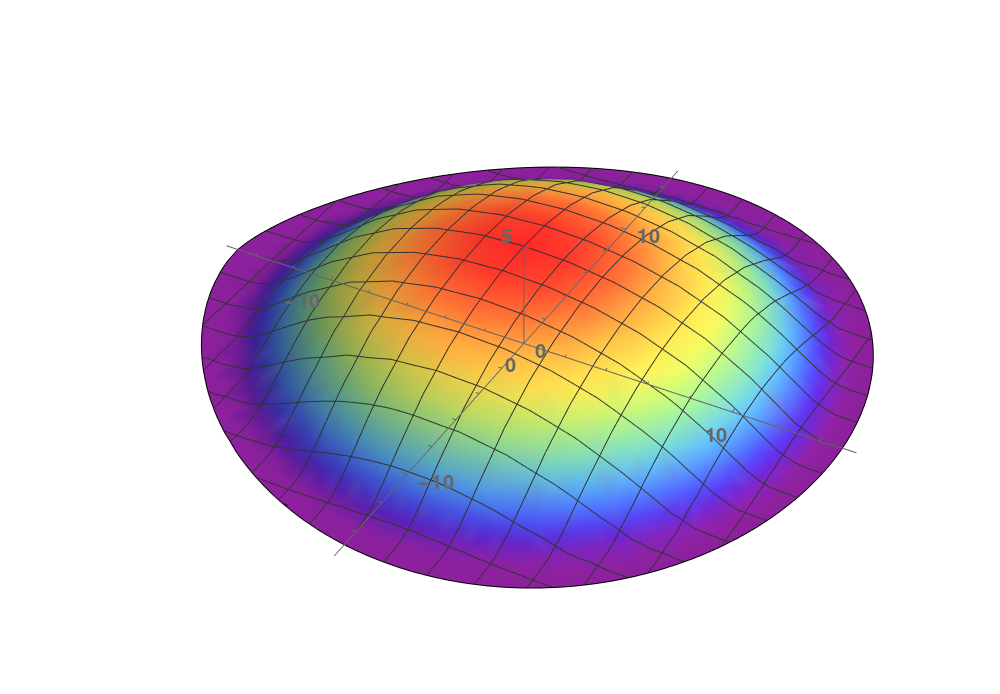}     &
\parbox{0.2\linewidth}{
 \begin{align*}
 l &= 30 \\
w &= 30 \\
d &= 5 \\
b &= e \\
p &= 1.2 \\
s_x &= 0 \\
s_y &= 0 \\
\end{align*}
} \\

\midrule
\includegraphics[width=\linewidth]{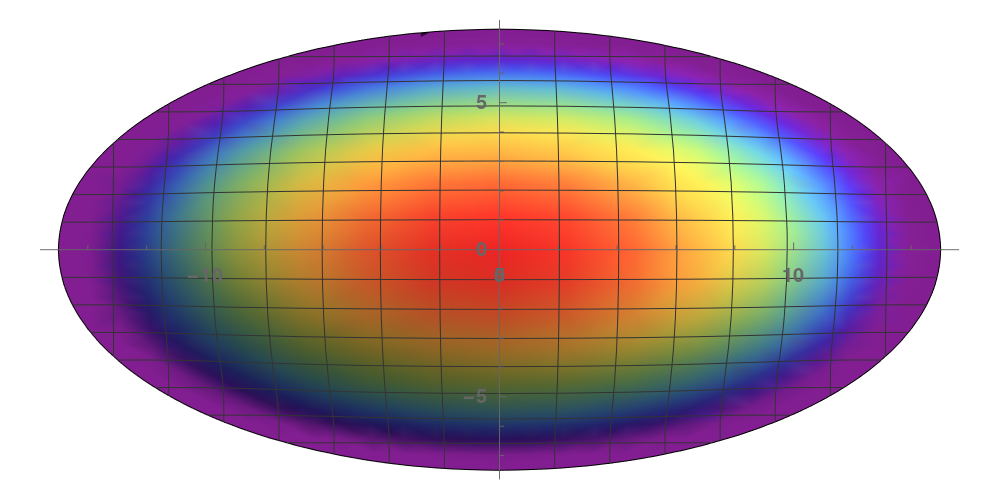} & \includegraphics[width=\linewidth]{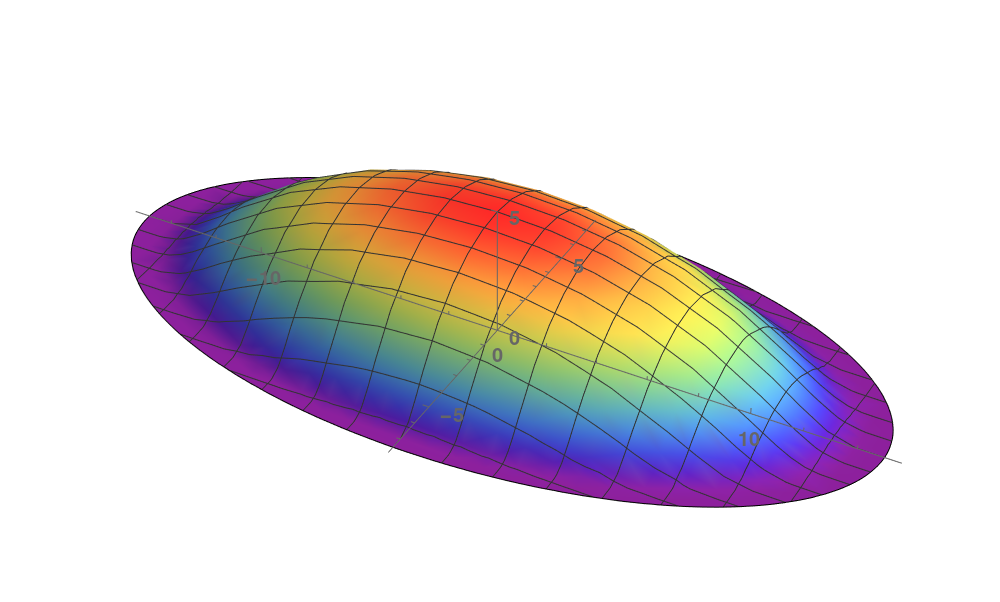}     &
\parbox{0.2\linewidth}{
 \begin{align*}
 l &= 30 \\
w &= 15 \\
d &= 5 \\
b &= e \\
p &= 1 \\
s_x &= 0 \\
s_y &= 0 \\
\end{align*}
} \\

\midrule
\includegraphics[width=\linewidth]{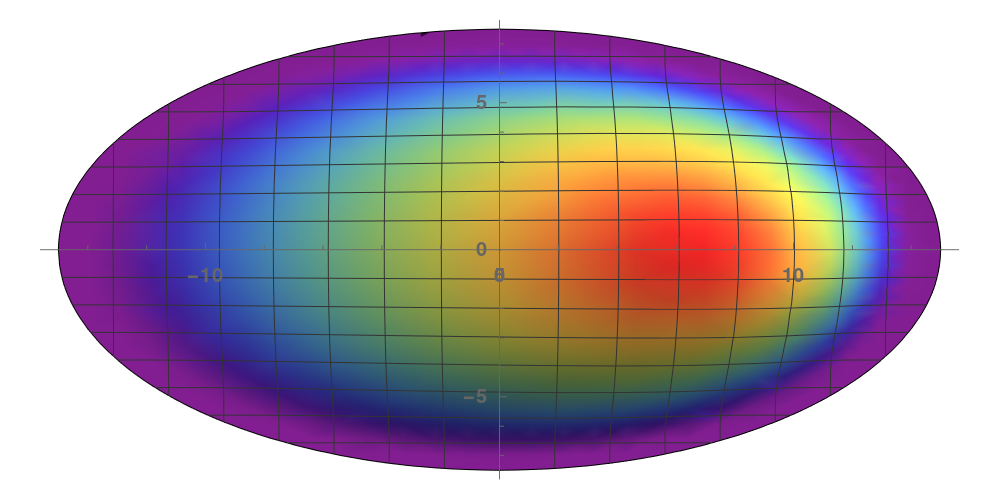} & \includegraphics[width=\linewidth]{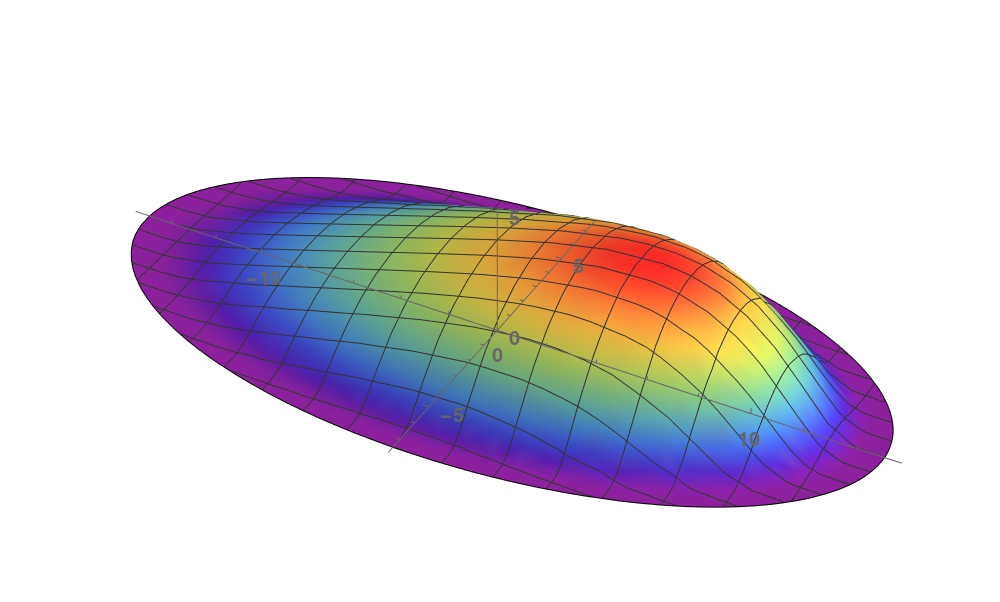}     &
\parbox{0.2\linewidth}{
 \begin{align*}
 l &= 30 \\
w &= 15 \\
d &= 5 \\
b &= e \\
p &= 1 \\
s_x &= 0.2 \\
s_y &= 0 \\
\end{align*}
} \\

\midrule
\includegraphics[width=\linewidth]{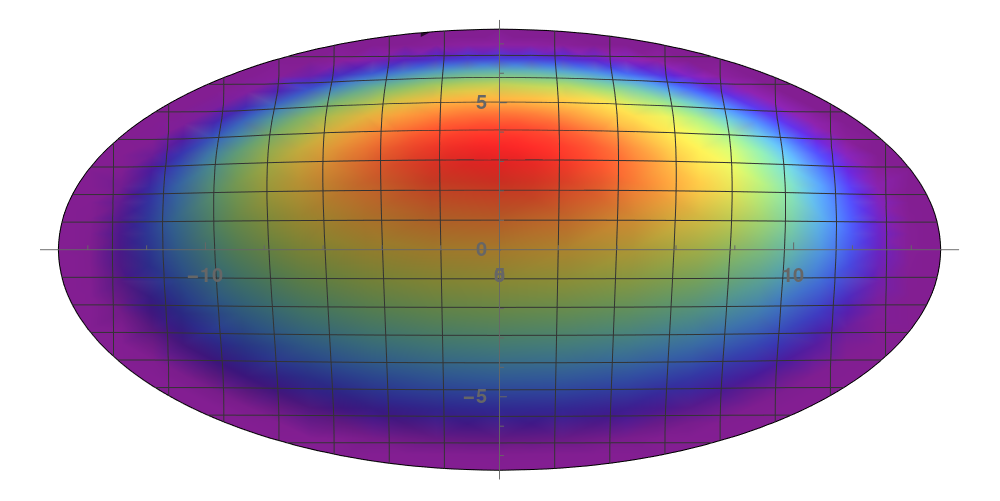} & \includegraphics[width=\linewidth]{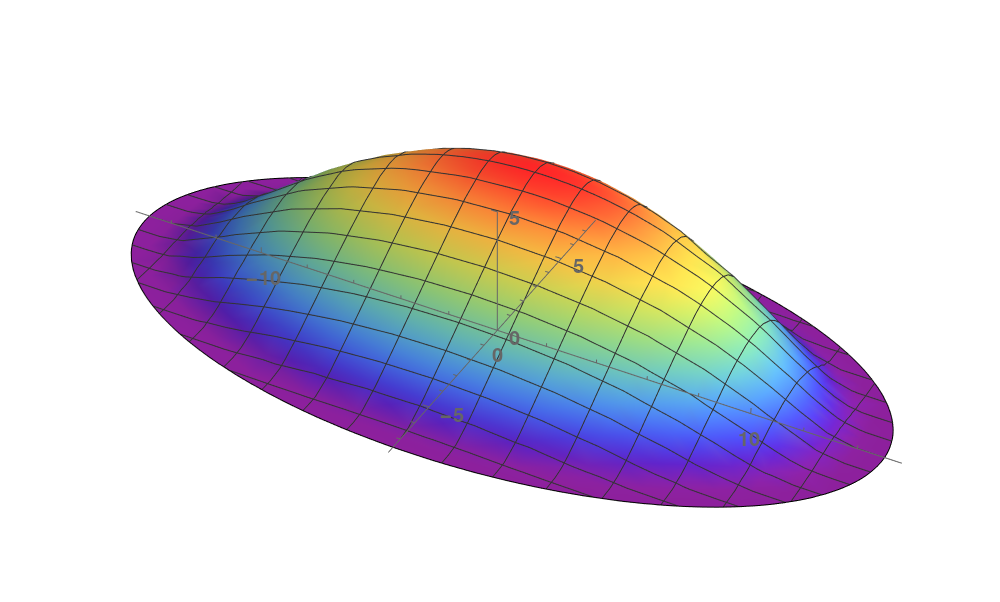}     &
\parbox{0.2\linewidth}{
 \begin{align*}
 l &= 30 \\
w &= 15 \\
d &= 5 \\
b &= e \\
p &= 1 \\
s_x &= 0 \\
s_y &= 0.2 \\
\end{align*}
} \\

\midrule
\includegraphics[width=\linewidth]{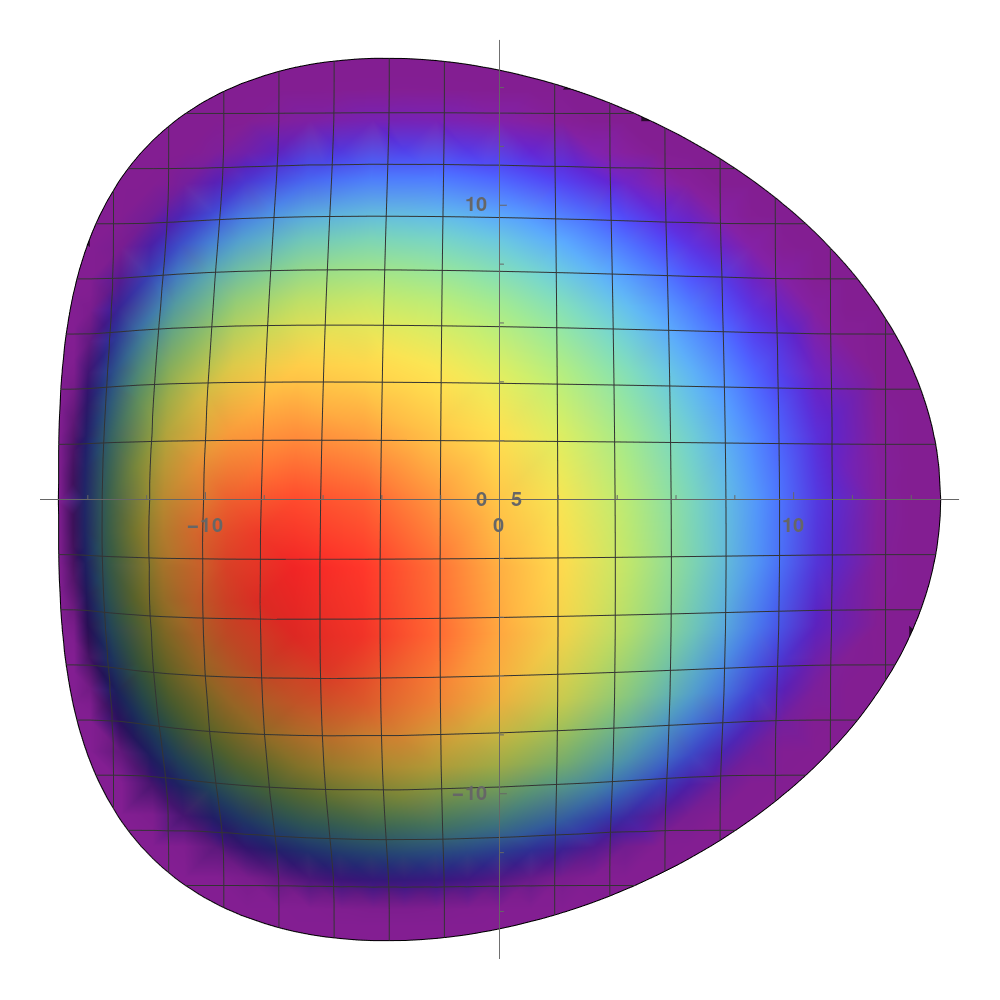} & \includegraphics[width=\linewidth]{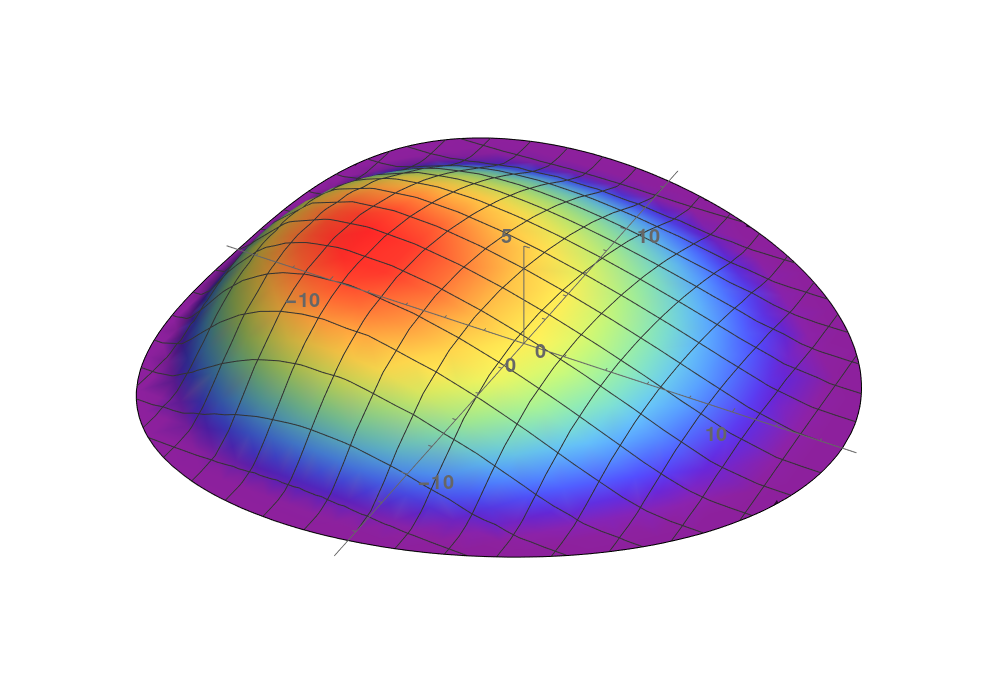}     &
\parbox{0.2\linewidth}{
 \begin{align*}
 l &= 30 \\
w &= 30 \\
d &= 5 \\
b &= e \\
p &= 0.7 \\
s_x &= -0.2 \\
s_y &= -0.1 \\
\end{align*}
} \\

\bottomrule
%\end{tabular}
\caption{Examples of the proposed dent model for different parameter values, where $e$ stands for the Euler's number.}
\label{tab:examples}
\end{longtable}
\end{center}

\section{INTENDED WORKFLOW}
\begin{figure}[h!]
 \centering
    \includegraphics[width=\linewidth]{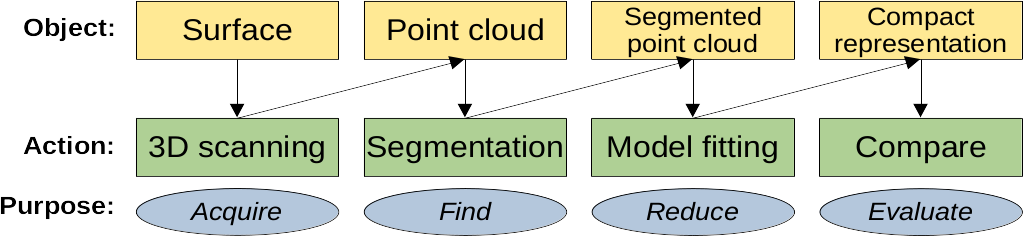}
  \caption{Expected workflow of an automated dent inspection system.}
  \label{fig:workflow}
\end{figure}

The model introduced in the previous section proposes the use of $7$ parameters that all have a physical meaning. However, they cannot be easily measured manually (e.g. with rules, gauges, etc.) and, thus, the model benefits from the combination with 3D scanning technologies. The intended workflow is displayed in Fig.~\ref{fig:workflow}.

Starting from the physical surface to be assessed, its point cloud representation is acquired. Then a \textit{segmentation} of some sort must be applied in order to actually isolate each dent by separating dented and non-dented areas~\cite{lafioscaDentSegmentation}.
In MRO, both the manual approach and the recent 3D scanners aim at extracting length, width and depth as prescribed by the SRM~\cite{reyno2018surface}.
Reducing the amount of information to a handful of parameters is necessary to make the measure intelligible and comparable.
However, as shown above, the sole measures of depth, width or length do not adequately represent the actual shape and create high degree of ambiguity in the following evaluation.

A more accurate alternative is available by means of the model proposed in Eq.~(\ref{eq:dent}), which can be fitted by numerical optimisation to the 3D points, providing the optimal set of $7$ parameters representing that particular dent shape.
This compact set of values will accurately represent the damage and will be \textit{compared} against historical data and structural evaluations, while such comparison would not be possible by directly using 3D point cloud data.
The added benefit is that storing or sharing the raw point cloud can be avoided, as the shape is faithfully represented by means of the proposed model.

The approach proposed in Fig.~\ref{fig:workflow} can lead to more targeted interventions, ultimately saving costs.
In order to accomplish this, the current practices are supposed to be revised and updated to embrace the increased capabilities brought by 3D scanning. This process of change has already started with the introduction of new devices, however, as part of a highly regulated and safety-critical industry, it is likely to take several years.

\section{USE CASES AND DISCUSSION}\label{sec:useCases}

\subsection{Report and comparison}
\begin{figure}[!ht]
 \centering
 \includegraphics[width=0.5\linewidth]{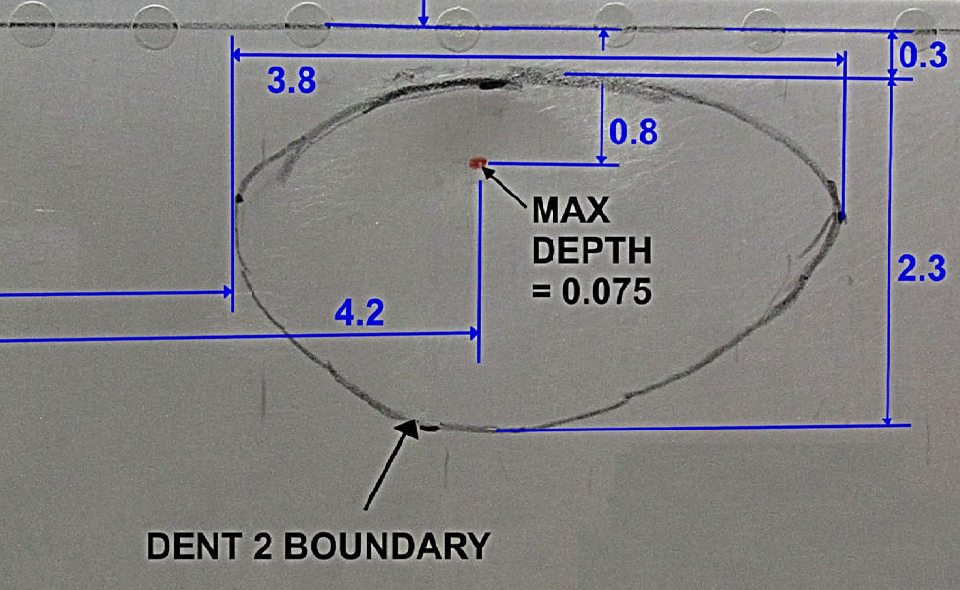}
 \caption{The complex shape of a dent cannot be expressed by only a combination of length, width and depth. In this picture the engineer relied on drawings to express the actual shape and the position of maximum depth, so that a better structural evaluation could be carried on. Even using 3D scanning, such great amount of data is ineffective if comparison is then done by means of depth and width/depth ratio only.}
\label{fig:pictureDentEvaluation}
\end{figure}

% Example of fitting! Images!
Fig.~\ref{fig:pictureDentEvaluation} is extracted from a ``C'' check report. It shows a dent manually marked and measured by the engineers. While this method may appear approximated and cumbersome, it responds to the need for communicating more about the actual dent shape, and thus obtain specific repair instructions from the aircraft manufacturer.
On the other hand, one may represent the damage by its full 3D point cloud. While this is rich of data, it is not easily comparable nor manageable in terms of extracting general repair recommendations.

\begin{figure}[h!]
 \centering
    \includegraphics[width=0.5\linewidth]{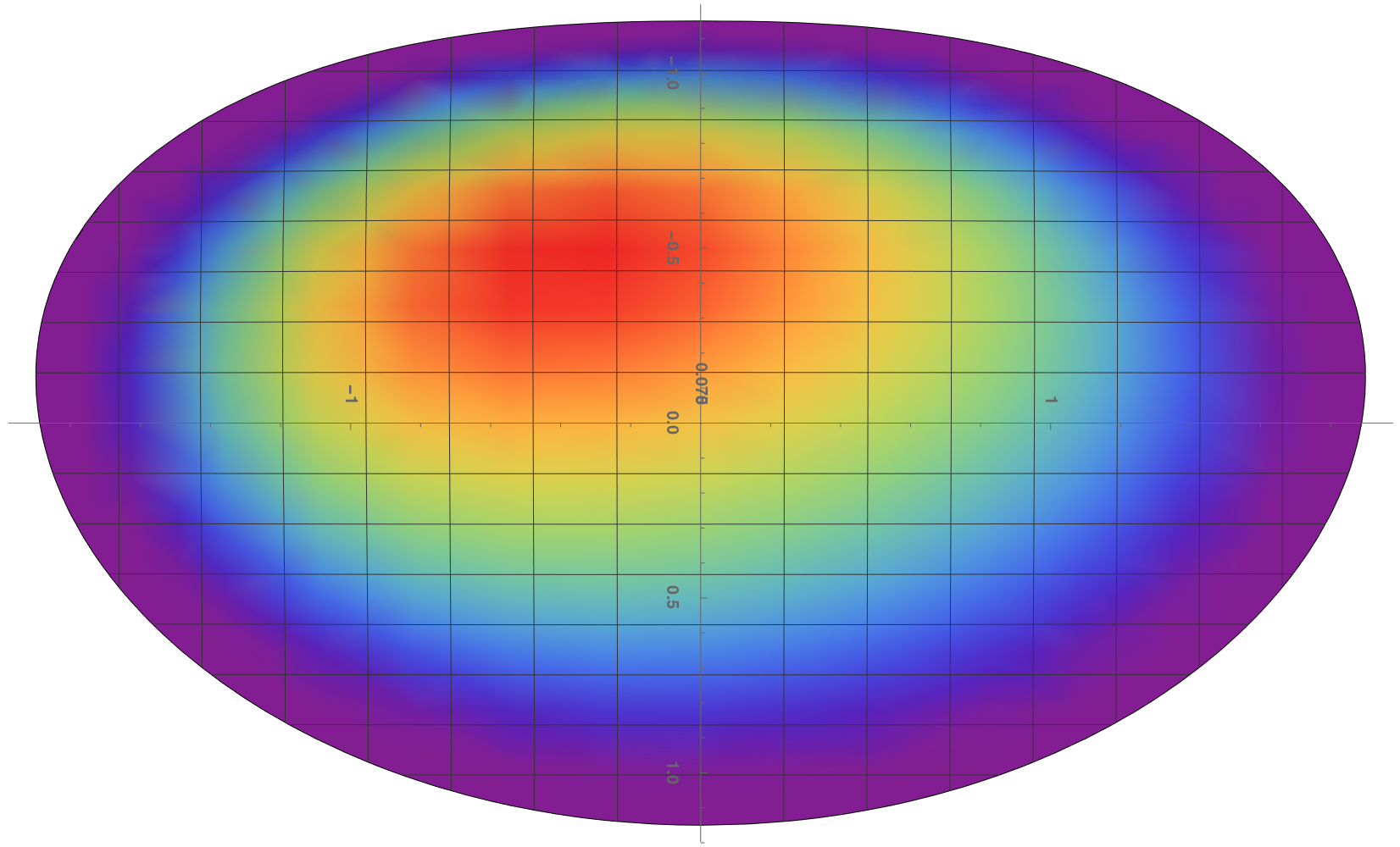}
  \caption{Shape obtained with $b=2.7$, $l=2.3$, $w=3.8$, $d=0.075$, $p=0.85$, $s_x=-0.2$ and $s_y=-0.12$ which could potentially represent the dent of Fig.~\ref{fig:pictureDentEvaluation}.}
  \label{fig:real_manual}
\end{figure}

Assuming the use of a 3D scanning, instead, the shape showed in Fig.~\ref{fig:real_manual}, calculated by means of Eq.~(\ref{eq:dent}), may easily approximate the actual one and with an intelligible set of only $7$ parameters. While length, width and depth are the same as in the classical evaluation, the $4$ extra degrees of freedom address the necessity of an accurate and compact description that was previously attempted by manual drawings only.

%The approach of using such model implicitly solves another problem:

% SRM: where maximum depth is measured? Usually where width is measured, but it is a simplification!!!

\subsection{Increased accuracy in the evaluation}
The increased accuracy of the model is hereafter shown by means of a 3D simulation in Blender.  Two different dents are obtained simulating a plane struck by a sphere with inclinations of $\SI{45}{\degree}$ and $\SI{60}{\degree}$ with respect to the normal of the plane. As the focus of this work is on the capability to represent complex shapes, the materials employed are not relevant.

To demonstrate the advantages introduced by the model, the two resulting dents were evaluated by a simplified 3 parameter model and the proposed model. Since traditional dent evaluation does not impose an exact shape, the simplified model was obtained by imposing $p = 1$, $s_x = s_y = 0$ and $b = \mathrm{e}$, leaving only $3$ degrees of freedom corresponding to length, width and depth.
It must be remarked that this is already an enhancement with respect to the traditional box model, however, the simulation showed a significant increase of accuracy, especially in the $\SI{60}{\degree}$ impact simulation, which justifies the introduction of the $7$-parameter model.
The mean average error (MAE) has here been used as comparison metric.

\begin{figure}[!h]
 \centering
 \begin{subfigure}[t]{0.3\linewidth}
 \centering
    \includegraphics[width=\linewidth]{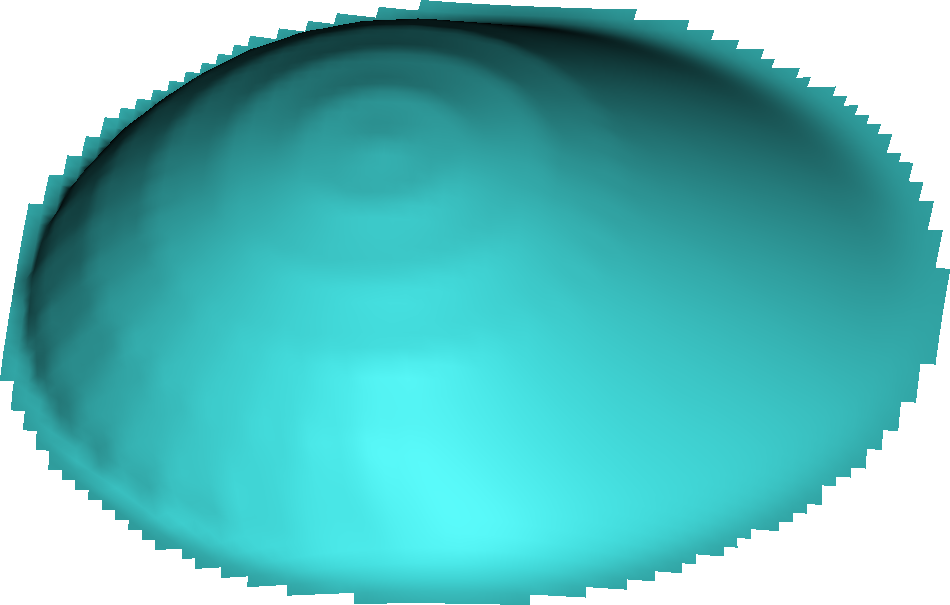}
    \caption{Original 3D dent shape.}
  \end{subfigure}
  \hfill
  \begin{subfigure}[t]{0.3\linewidth}
  \centering
    \includegraphics[width=\linewidth]{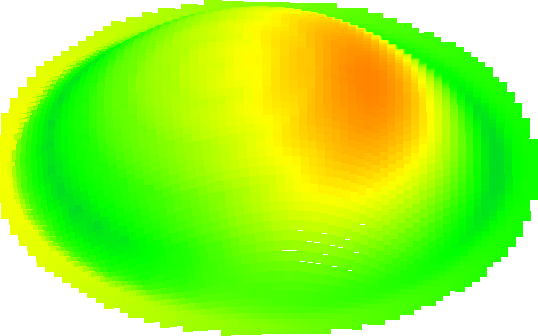}
    \caption{Simplified 3-parameter fitting and heatmap of residuals.}
  \end{subfigure}
  \hfill
  \begin{subfigure}[t]{0.3\linewidth}
  \centering
    \includegraphics[width=\linewidth]{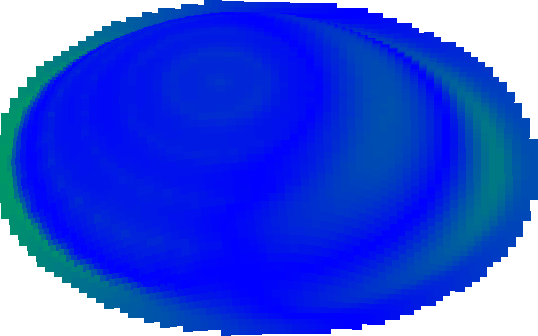}
    \caption{Proposed $7$-parameter fitting and heatmap of residuals.}
  \end{subfigure}
  %\caption*{\rightline{\scriptsize{Credits: .} }}
  \caption{Dent obtained from a $\SI{45}{\degree}$ collision and residual heatmaps (blue = $\SI{0}{\milli\meter}$, red = $\SI{1}{\milli\meter}$ mm). The proposed model is capable to follow the natural dent shape and the MAE is reduced by an order of magnitude.}
  \label{fig:cloth45All}
\end{figure}

% Maximum depth from 3D model is 1.35
In the $\SI{45}{\degree}$ collision example, the fitting by the simplified model gave $l = \SI{5.55}{\milli\meter}$, $w = \SI{5.09}{\milli\meter}$ and $d = \SI{1.46}{\milli\meter}$ with a MAE of $\SI{0.51}{\milli\meter}$.
The proposed model found $l = \SI{6.10}{\milli\meter}$, $w = \SI{5.48}{\milli\meter}$, $ d = \SI{1.24}{\milli\meter}$, $ p = \SI{1.01}{}$, $ s_x = \SI{-0.11}{}$, $ s_y = \SI{0.00}{}$ and $ b = \SI{3.11}{}$, with a MAE of only $\SI{0.053}{\milli\meter}$. The original 3D dent and the two residuals are shown in Fig.~\ref{fig:cloth45All}, where red indicates high residual areas.

\begin{figure}[!h]
 \centering
 \begin{subfigure}[t]{0.3\linewidth}
 \centering
    \includegraphics[width=\linewidth]{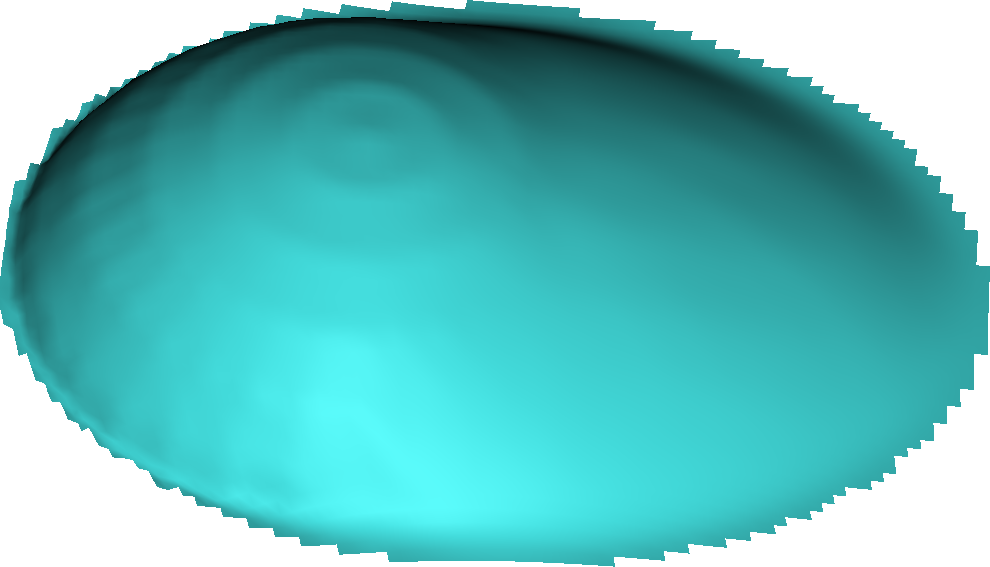}
    \caption{Original 3D dent shape.}
  \end{subfigure}
  \hfill
  \begin{subfigure}[t]{0.3\linewidth}
  \centering
    \includegraphics[width=\linewidth]{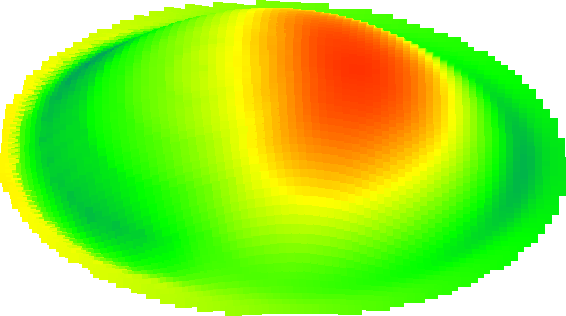}
    \caption{Simplified 3-parameter fitting and heatmap of residuals.}
  \end{subfigure}
  \hfill
  \begin{subfigure}[t]{0.3\linewidth}
  \centering
    \includegraphics[width=\linewidth]{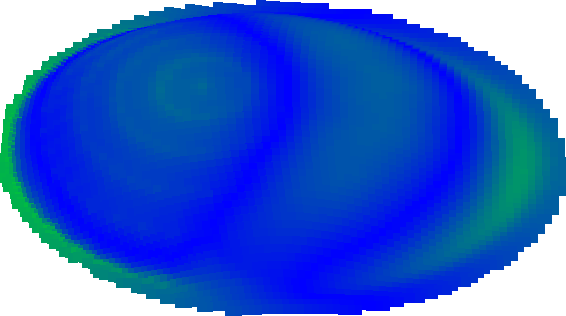}
    \caption{Proposed $7$-parameter fitting and heatmap of residuals.}
  \end{subfigure}
  %\caption*{\rightline{\scriptsize{Credits: .} }}
  \caption{Dent obtained from a $\SI{60}{\degree}$ collision and residual heatmaps (blue = $\SI{0}{\milli\meter}$, red = $\SI{1}{\milli\meter}$ mm). Also in this case the MAE is greatly reduced.}
  \label{fig:cloth60All}
\end{figure}

% Maximum depth from 3D model is 1.35
In the $\SI{60}{\degree}$ collision example, the fitting by the simplified model gave $l = \SI{6.36}{\milli\meter}$, $w = \SI{5.14}{\milli\meter}$ and $d = \SI{1.48}{\milli\meter}$ with a MAE of $\SI{0.52}{\milli\meter}$.
The proposed model found $l = \SI{6.66}{\milli\meter}$, $w = \SI{5.21}{\milli\meter}$, $ d = \SI{1.12}{\milli\meter}$, $ p = \SI{0.96}{}$, $ s_x = \SI{-0.12}{}$, $ s_y = \SI{0.00}{}$ and $ b = \SI{2.37}{}$, with a MAE of only $\SI{0.08}{\milli\meter}$. Fig.~\ref{fig:cloth60All} represents the original shape and the fitting residuals.

Using parameters that allow to represent the actual shape is an advantage in terms of evaluation reliability, as residual strength is influenced by the many characteristics of the damage, not limited to material or depth~\cite{li2015effect,jing2021inspection}.
Moreover, the above examples show that the $7$ parameters are able to reproduce the common dent shape with great accuracy, reducing the MAE by an order of magnitude.

\subsection{Application to real dents}\label{subsec:8tree}
\begin{figure}[h!]
 \centering
    \includegraphics[width=0.5\linewidth]{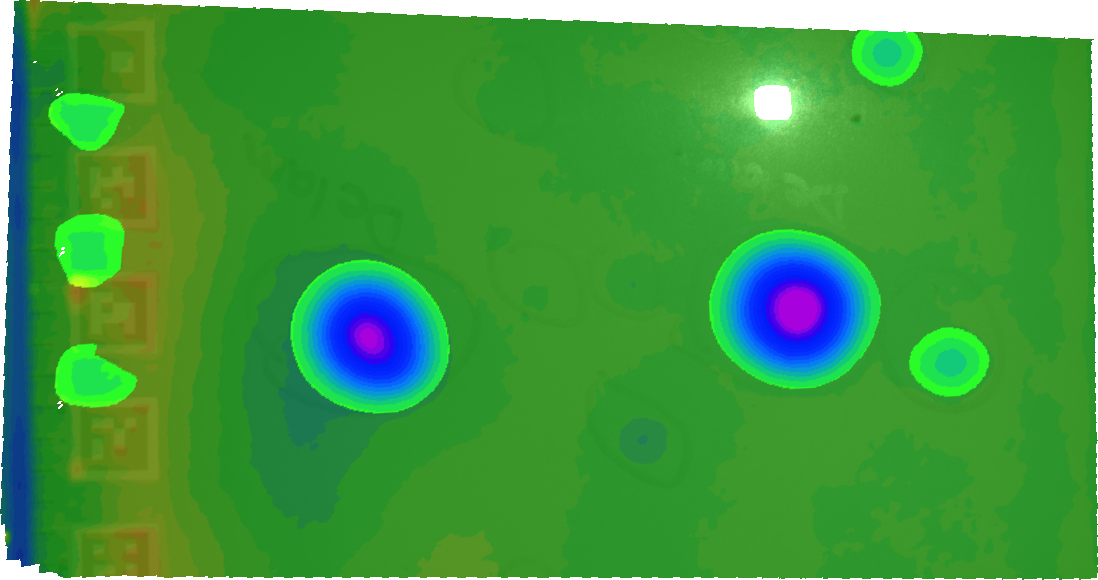}
  \caption{One of the point clouds from which dents were extracted.}
  \label{fig:20211203T213430}
\end{figure}

In order to validate the approach, the $7$ parameters evaluation was applied to point clouds obtained by 8tree’s \textit{dentCHECK} tool. Fig.~\ref{fig:20211203T213430} shows one of the point clouds from which the dents where isolated.

\begin{figure}[!h]
 \centering
 \begin{subfigure}[t]{0.3\linewidth}
 \centering
    \includegraphics[width=\linewidth]{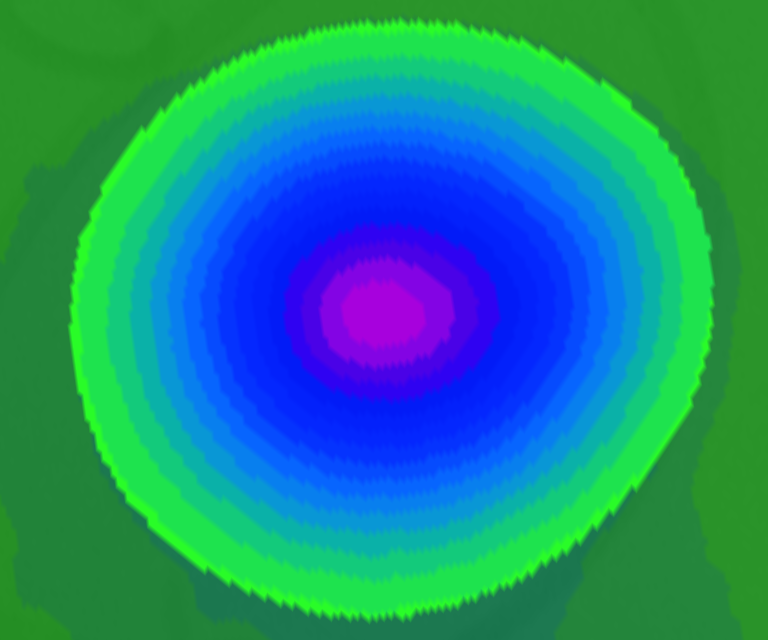}
    \caption{Dent as scanned with dentCHECK.}
  \end{subfigure}
  \hfill
  \begin{subfigure}[t]{0.3\linewidth}
  \centering
    \includegraphics[width=0.9\linewidth]{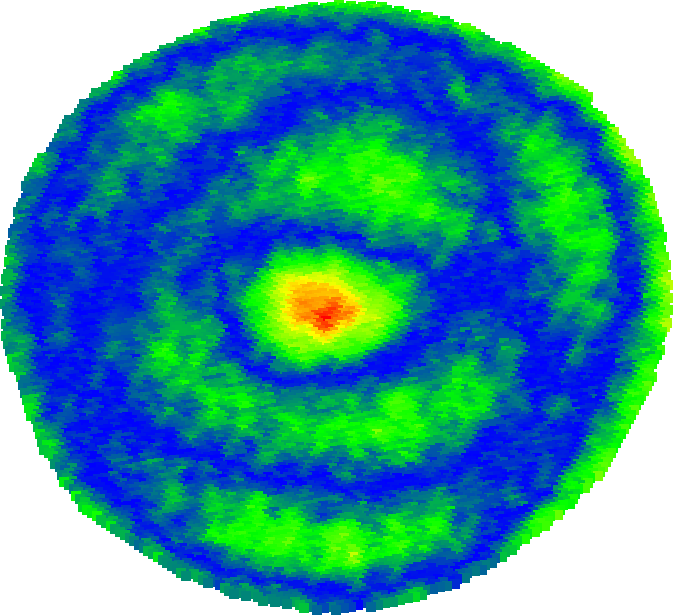}
    \caption{Heatmap of residuals.}
  \end{subfigure}
  \hfill
  \begin{subfigure}[t]{0.3\linewidth}
  \centering
    \includegraphics[width=\linewidth]{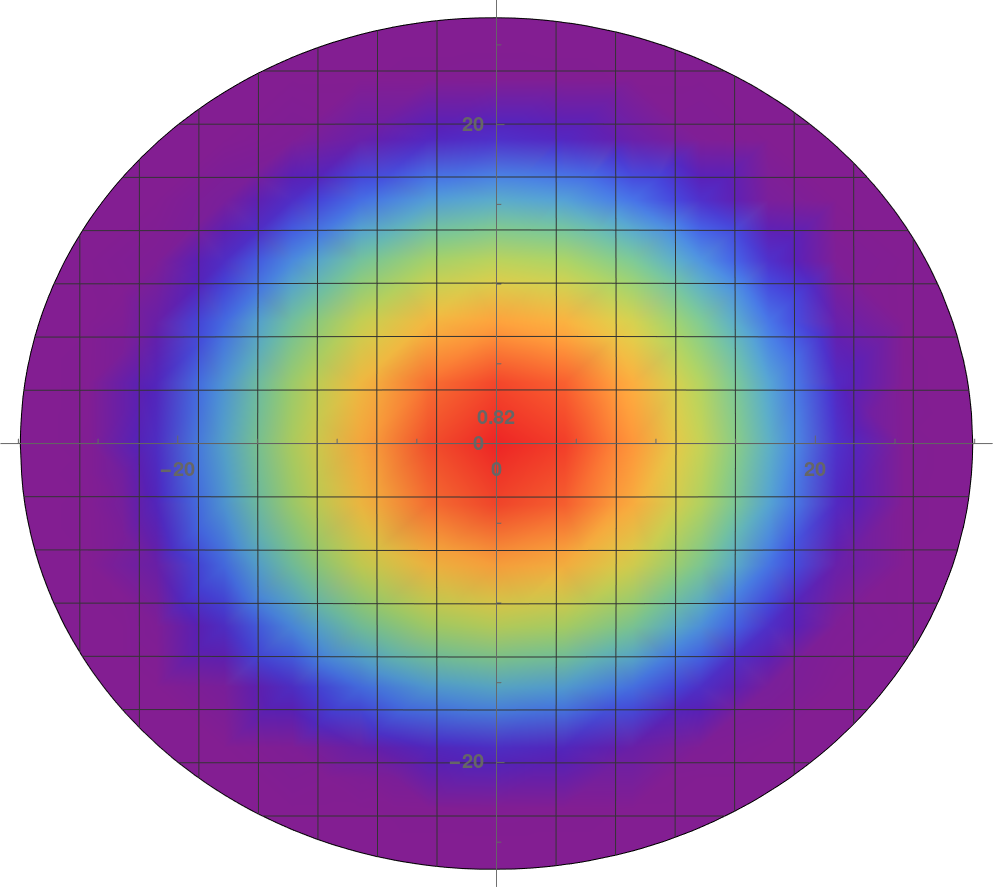}
    \caption{Fitted model.}
  \end{subfigure}
  \hfill
  %\caption*{\rightline{\scriptsize{Credits: .} }}
  \caption{Evaluation on the first dent. The MAE was $\SI{0.02}{\milli\meter}$.}
  \label{fig:20211203T213430_dent1}
\end{figure}

After fitting the model, the first dent had $l = \SI{59.76}{\milli\meter}$, $w = \SI{53.42}{\milli\meter}$, $ d = \SI{0.82}{\milli\meter}$, $ p = \SI{0.99}{}$, $ s_x = \SI{0.01}{}$, $ s_y = \SI{0.00}{}$ and $ b = \SI{10.00}{}$, with a MAE of $\SI{0.02}{\milli\meter}$. The isolated dent and its corresponding model are shown in Fig.~\ref{fig:20211203T213430_dent1}.

\begin{figure}[!h]
 \centering
 \begin{subfigure}[t]{0.3\linewidth}
 \centering
    \includegraphics[width=\linewidth]{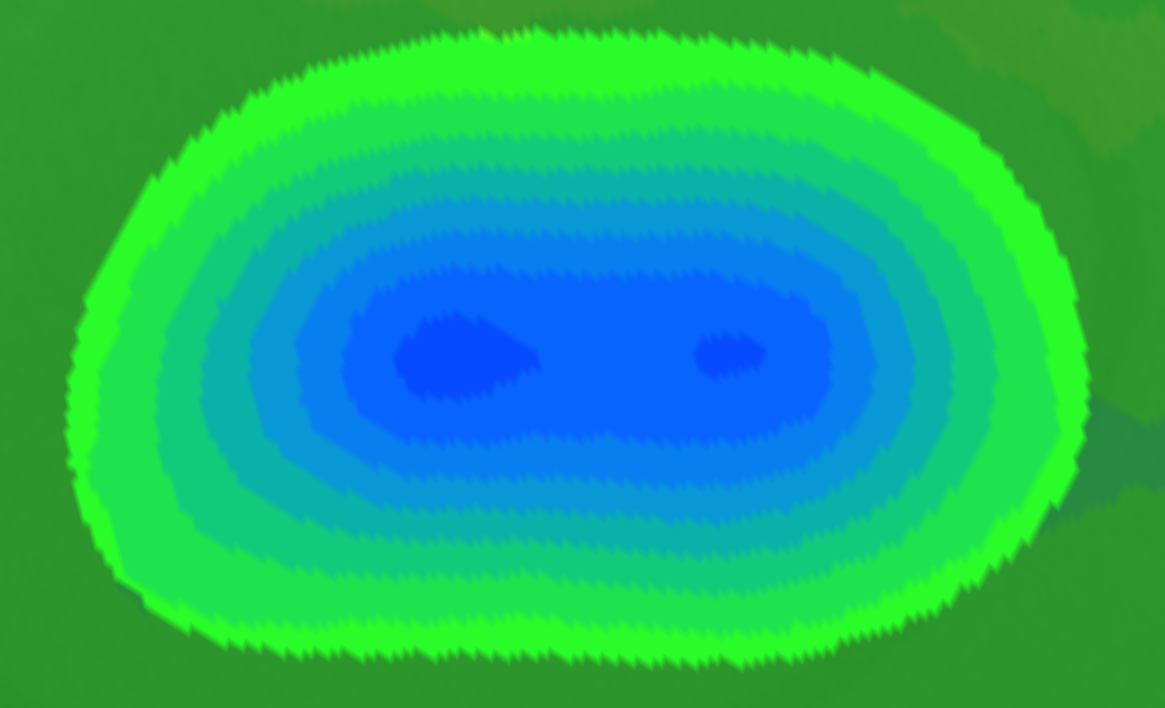}
    \caption{Dent as scanned with dentCHECK.}
  \end{subfigure}
  \hfill
  \begin{subfigure}[t]{0.3\linewidth}
  \centering
    \includegraphics[width=0.9\linewidth]{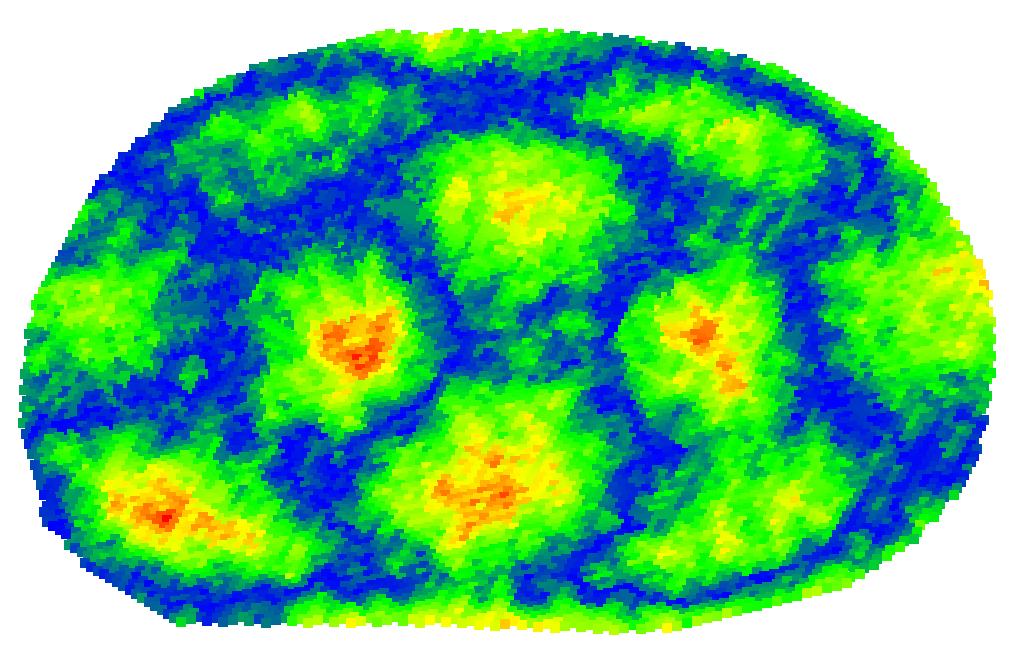}
    \caption{Heatmap of residuals.}
  \end{subfigure}
  \hfill
  \begin{subfigure}[t]{0.3\linewidth}
  \centering
    \includegraphics[width=\linewidth]{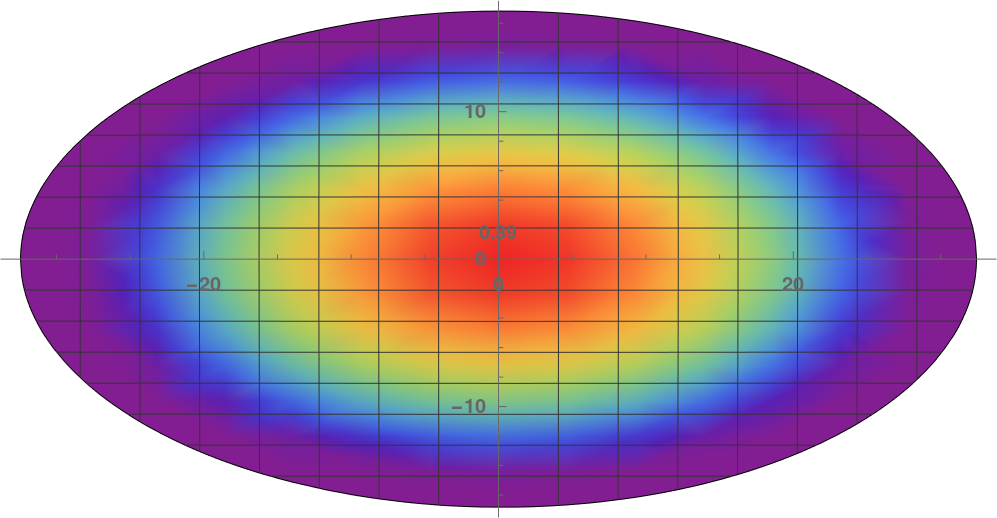}
    \caption{Fitted model.}
  \end{subfigure}
  \hfill
  %\caption*{\rightline{\scriptsize{Credits: .} }}
  \caption{Evaluation on the second dent. The MAE was $\SI{0.02}{\milli\meter}$.}
  \label{fig:20211203T213545_dent1}
\end{figure}

% 20211203T213545_dent1
The same was repeated on a second, more irregular dent, producing $l = \SI{64.88}{\milli\meter}$, $w = \SI{33.66}{\milli\meter}$, $ d = \SI{0.39}{\milli\meter}$, $ p = \SI{1.01}{}$, $ s_x = \SI{0.01}{}$, $ s_y = \SI{0.00}{}$ and $ b = \SI{4.64}{}$, with a MAE of $\SI{0.02}{\milli\meter}$ but with less accuracy in representing the edge. Details are shown in Fig.~\ref{fig:20211203T213545_dent1}.

Such approach shows how a compact 3D representation of the damage can be efficiently provided.
Nevertheless, limitations to be addressed comprise, for example, cases when multiple overlapping dents are present or when the original (undamaged) surface is curved. To address these, a more advanced version of the optimisation algorithm may be developed, thus finding the best match in the different conditions.

\section{CONCLUSIONS AND FUTURE WORK}\label{sec:DentConclusions}

The MRO industry is gradually embracing 3D scanning devices for dent inspections, capable to easily acquire complex shapes. However, the accuracy and the amount of information that such devices offer is deeply jeopardised by the traditional dent measuring methodology, which models the dent as a ``box'', i.e. only by their basic measures and without taking its actual shape into account.
Examples showed how the same length, width and depth may actually correspond to very different dent shapes, undermining the objective assessment of the engineer. For asymmetric dents, often the measured depth may not match the maximum damage depth.
On the other hand, a full point cloud or a drawing representing the dent are not readily intelligible nor easily comparable, because they cannot be given as a compact set of parameters.

The model proposed in this work enables the representation of complex dent shapes by adding $4$ dimensional parameters ($b$, $p$, $s_x$ and $s_y$) to the traditional $3$ defined in the SRM, namely length, width and depth.
The model proposes a compact and convenient way to describe dent damage that may pave the way for a novel dent definition in the SRM and an efficient reformulation of allowable damage thresholds. This, in turn, would allow more accurate reporting, more targeted repairs and, ultimately, reduced costs. Computer physics simulations showed that the mean average error is reduced by an order of magnitude when using the proposed model. Test on real scan data showed how the model can be applied to real damages.

In future, the residual strength of the skin should be related to different dent parameters and materials. The model could be then integrated in commercial MRO 3D scanning systems to provide the optimal parameters via advanced numerical optimisation, leading to a new reliable way to evaluate this very common damage.

%\vspace*{-15pt}
\section*{ACKNOWLEDGEMENTS}
The authors would like to thank the company 8tree for providing data from their dentCHECK tool, whose point clouds have been used for the validation of Sec.~\ref{subsec:8tree}.

\section*{DECLARATIONS}
All authors certify that they have no affiliations with or involvement in any organization or entity with any financial interest or non-financial interest in the subject matter or materials discussed in this manuscript. The authors did not receive support from any organization for the submitted work.

Data used in this article is subject to third party restrictions.

%\cleardoublepage
\printbibliography

\end{document}